\documentclass{article}

\usepackage[round]{natbib}

\usepackage{amsmath, amssymb, graphicx}
\usepackage[margin=1in]{geometry}
\usepackage[utf8x]{inputenc}
\usepackage{algorithm}
\usepackage[noend]{algpseudocode}
\usepackage{comment}
\usepackage{scrextend}
\usepackage[usenames, dvipsnames]{color}
\usepackage{bbm}
\usepackage{mathtools}
\usepackage{comment}
\usepackage[table,xcdraw]{xcolor}

\newcommand\ta{\hat{\theta}_{\text{avg}}}
\newcommand\ts{\hat{\theta}_{\text{shape}}}
\newcommand\tf{\hat{\theta}_{\text{frac}}}
\newcommand\CSC{C_\text{SC}}
\newcommand\cSC{c_\text{SC}}
\newcommand\SSC{S_\text{SC}}
\newcommand\ch{\hat{\chi}} 
\newcommand\cbh{\hat{\bar{\chi}}} 
\newlength{\hcolw} 
\DeclarePairedDelimiter\ceil{\lceil}{\rceil}
\DeclarePairedDelimiter\floor{\lfloor}{\rfloor}

\newcommand\tsn{\hat{\theta}_{\text{s}}^n}
\newcommand\tfn{\hat{\theta}_{\text{f}}^n}
\newcommand\nf{\floor{np/2}}

\twocolumn
\begin{document}

\title{Copula Quadrant Similarity for Anomaly Scores}
\author{Davidow, Matthew \and  Matteson, David S. }
\maketitle

\begin{abstract}
  Practical anomaly detection requires applying numerous approaches due to the inherent difficulty of unsupervised learning. Direct comparison between complex or opaque anomaly detection algorithms is intractable; we instead propose a framework for associating the scores of multiple methods. Our aim is to answer the question: how should one measure the similarity between anomaly scores generated by different methods? The scoring crux is the extremes, which identify the most anomalous observations. A pair of algorithms are defined here to be similar if they assign their highest scores to roughly the same small fraction of observations. To formalize this, we propose a measure based on extremal similarity in scoring distributions through a novel upper quadrant modeling approach, and contrast it with tail and other dependence measures. We illustrate our method with simulated and real experiments, applying spectral methods to cluster multiple anomaly detection methods and to contrast our similarity measure with others. We demonstrate that our method is able to detect the clusters of anomaly detection algorithms to achieve an accurate and robust ensemble algorithm.
\end{abstract}
\section{Introduction}
Unsupervised anomaly detection (no labeled data for training) is an especially hard problem because anomalies are ambiguously defined. As such, ensemble approaches are commonly utilized \citep{aggarwal2015outlier}. We aim to understand the similarity and dependence among scores from multiple anomaly detection algorithms. 
This is necessary to better understand which algorithms tightly cluster their largest scores (and are largely redundant) and which have complementary strengths in identifying different anomalies.
Quantifying similarity is also necessary to optimally design an ensemble of diverse anomaly detection scoring methods \citep{jaffe2016unsupervised}. 

Simple correlation has limited use here, what is of interest is the association between extreme scores, which correspond to the relatively most anomalous observations as inferred by each algorithm. 
We propose a novel upper quadrant based similarity measure that is sensitive to upper tail dependence 
through a copula distribution. This similarity measure averages a new 
copula quadrant maximum likelihood estimator 
and a survival copula estimator, inheriting the strengths of both with minimal compromise. This average is robust in practice, which is necessary considering the varied and ambiguous nature of anomalies.


Extremal dependence measures are also important in related applications: 
empirical finance \citep{caillault2005empirical}, 
medical insurance claims \citep{cebrian2003analysis},
and weather extremes \citep{serinaldi2008analysis}.
Copula models have had success in modeling data with frequent co-extremes \citep{juri2002copula}, and we build on this work but with additional focus on non-limiting tail dependence through the upper quadrant models defined below. 


%
\paragraph{Related Methods.} 
There are many existing measures for extremal dependence \citep{beirlant2006statistics},  however, most are either distributionally holistic and not sufficiently sensitive specifically to tail dependence, or focus specifically on the limiting tail behavior as the quantiles approach $1$. However we are interested in the more practical regime, where the expected anomaly fraction may range from 0 - 15\%, for different applications.
\cite{chang2016robust} defines an overall dependence measure with respect to the L1 distance between a two dimensional copula and the uniform copula, however this gives little emphasis to tail dependence.

\begin{figure}
\begin{centering}
\includegraphics[width=0.22\textwidth]{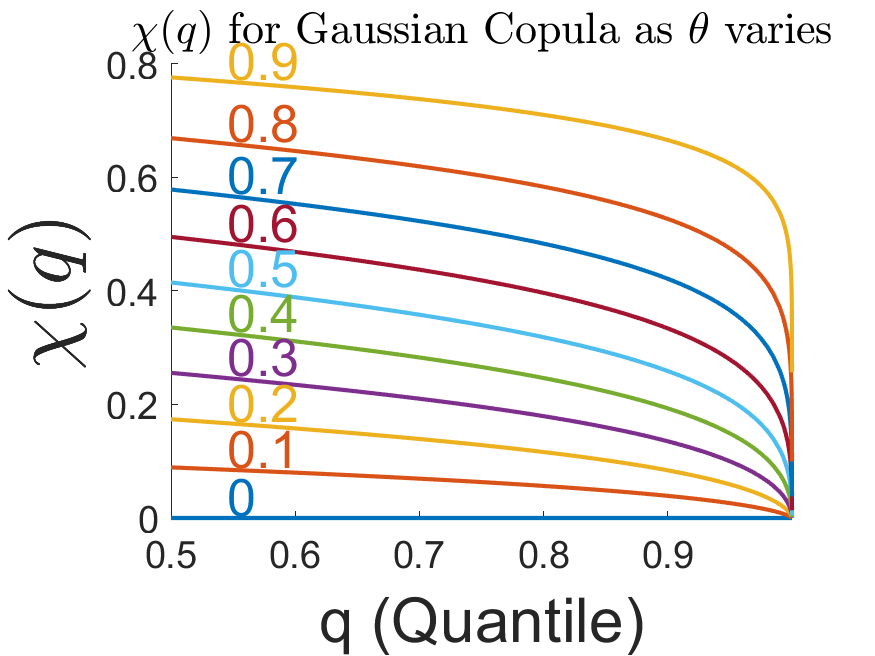}
\includegraphics[width=0.22\textwidth]{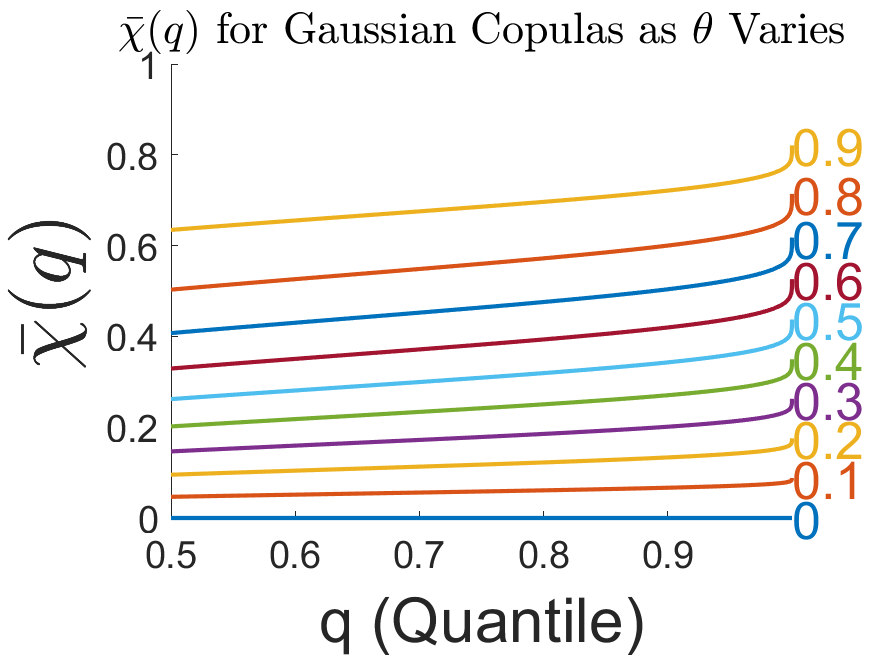}
\includegraphics[width=0.22\textwidth]{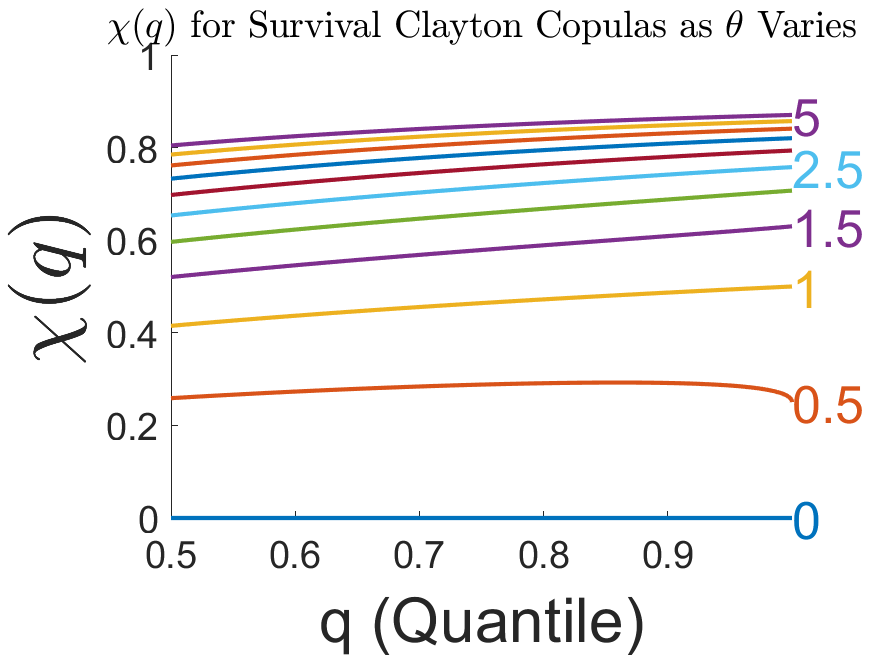}
\includegraphics[width=0.22\textwidth]{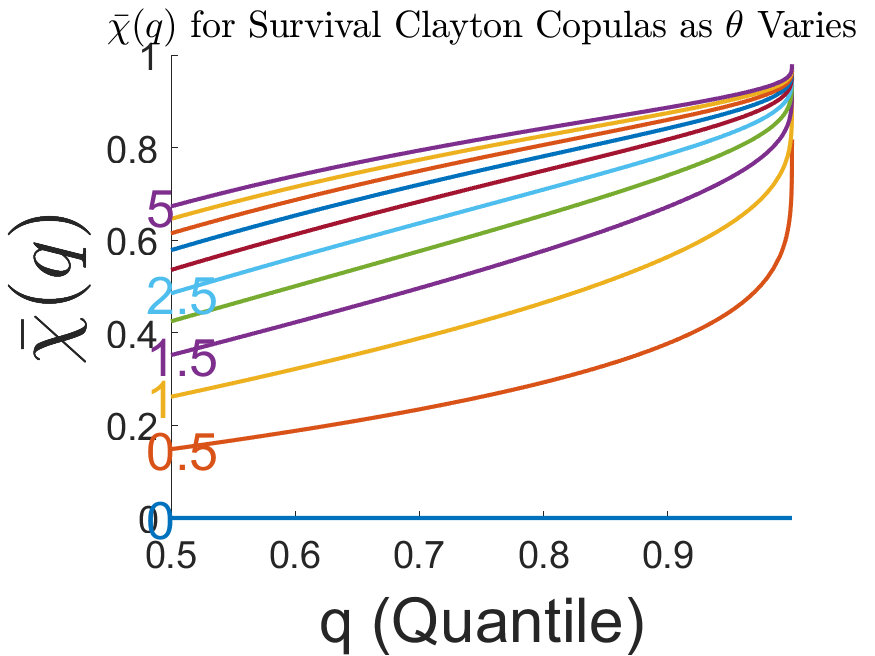}
\caption{\label{fig:XiBar} The $\chi$ similarity measure (left) and $\bar{\chi}$ (right) for various association parameters $\theta$ for the Gaussian copulas (top) and survival Clayton copulas (bottom). 
The labeled lines each correspond to a specific (copula) distribution, with increasing values of $\theta$ corresponding to greater similarity. Note the Gaussian copula possess limiting tail independence with $\chi = 0$, whereas $\bar{\chi}$ still distinguishes different associations $\theta$.}
\end{centering}
\end{figure}

A natural tail dependence measure for a pair of random variables $(Y_1,Y_2)$ with continuous distributions and marginals $F_{Y_1}$,$F_{Y_2}$ 
is given by $\chi = \lim_{q \uparrow 1} \chi(q)$,
in which 
\begin{equation}\label{eq:Chi} 
\begin{split}\chi(q) &= P(F_{Y_1}(Y_1) > q | F_{Y_2}(Y_2) >q)  \\
&= P(U_1 > q| U_2 >q), 
\end{split}
\end{equation}
where $U_i = F_{Y_i}(Y_i)$ denotes the probability integral transformed marginal components. 
This (copula) \emph{upper tail limit} is sometimes denoted as $\lambda_U$, but $\chi(q)$ itself may also be taken as an dependence measure at any fixed $q \in (0,1)$.
\cite{coles1999dependence} builds upon work by \cite{joe1997multivariate} and proposes an alternative dependence measure, $\bar{\chi} = \lim_{q \uparrow 1} \bar\chi(q)$, in which 
\begin{equation}\label{eq:ChiBar}  \bar{\chi}(q) = 2\frac{\text{log} \, P(U_1>q)}{\text{log} \, P(U_1>q,U_2>q)} - 1.  \end{equation}
 This dependence measure is similar to $\chi$ except in the limiting behavior as $q \rightarrow 1$, as shown in figure \ref{fig:XiBar} with respect to Gaussian and Clayton copulas (defined below) with various association parameters.   
Sample estimates $\hat{\chi}(q)$ and $\hat{\bar{\chi}}(q)$ are defined by equations 
\ref{eq:Chi} and \ref{eq:ChiBar} using empirical probabilities.

\paragraph{Our Approach.}
We propose an upper quadrant based similarity measure tailored for comparing anomaly scores.
Our upper quadrant measure is novel with excellent practical performance, and allows high sensitivity to extremal dependence while being agnostic to the scale and distribution differences commonly found between the scores of diverse anomaly detection algorithms.
We introduce both model-based and empirical upper quadrant measures and find that an average of both is best across a wide range of anomaly scenarios and experiments. 

 
Our focus on only joint upper quantiles is especially suited to the analysis of anomaly scores obtained from diverse methods, as many detection algorithms indiscriminately rank inliers among the lower quantiles of scores, whereas the association of outliers' scores is the primary interest in comparing anomaly techniques. We thus derive similarity measures and define our upper quadrant model based on quantifying dependence withhin the portion of distribution of higher quantiles.
%
%
Our main contributions include: (i) a novel similarity measure sensitive to tail dependence, emphasizing practical usage; (ii) a novel metric for comparing competing similarity measures, (iii) an empirical comparison of similarity measures applied to anomaly scoring methods across a variety of data. 
%

\section{Copula Definitions}
%
A copula $C(\cdot)$ is a multivariate distribution (cumulative density function, or CDF) in which all univariate marginal distributions are standard uniform distributed on $[0,1]$. For a $d$-dimensinoal random variable $Y = [Y_1, \ldots , Y_d]$ with continuous and strictly monotonically increasing marginal cdfs $F_{Y_1},\ldots,F_{Y_d}$, the transformed random variable $U = [F_{Y_1}(Y_1),\ldots,F_{Y_d}(Y_d)]$ has a CDF which is also a copula, since each $F_{Y_i}(Y_i)$ is uniformly distributed. The copula function associated with $Y$ is denoted $C_Y(\mathbf{u})$ and the corresponding density is denoted as $c_Y(\mathbf{u}) = \frac{\partial C^d_Y(u_1,\ldots,u_d)}{\partial u_1 \ldots \partial u_d}$.   
In fact any multivariate distribution can be written in terms of its copula function as described by Sklar's Theorem \citep{ruschendorf2009distributional}: 
\begin{equation}\begin{split} \label{eq:Sklar}
& F_Y(y_1,\ldots,y_d) = C_Y(F_{Y_1}(y_1),\ldots,F_{Y_d}(y_d)) \\
&C_Y(u_1,\ldots,u_d) = F_Y(F_{Y_1}^{-1}(u_1),\ldots,F_{Y_d}^{-1}(u_d)).
\end{split} \end{equation}
%
%
%
These relations illustrate the central properties of copulas; they can be used to separate the marginal distribution from the dependence structure between variables. That is, the copula $C_Y$ contains all of the information about the dependencies between the variables $\{Y_i\}$, agnostic to the distributions $\{F_{Y_i}\}$.

\subsection{Copula Families}
There exists a rich literature on the theory of copulas, and many parametric families of copulas have been proposed \citep{ruppert2011statistics,joe1997multivariate}. Most commonly used is the Gaussian copula, shown in equation \ref{eq:GaussCop}, although moving forward we will focus on the Clayton copula shown in equation \ref{eq:ClayCop}.
The bivariate Gaussian copula is defined as
\begin{equation}\label{eq:GaussCop} 
C_{\text{Gauss}}(u_1,u_2| \theta) = \Phi_\theta (\Phi^{-1}(u_1),\Phi^{-1}(u_2)).
\end{equation}
Where $\Phi$ is the CDF of a standard univariate normal, and $\Phi_\theta$ is the bivariate CDF of a normal with unit variances and correlation (association) parameter $\theta \in (-1,1)$. 
The bivariate Clayton copula with association parameter $\theta \in (0, \infty)$ is defined as 
\begin{equation}\label{eq:ClayCop} 
C_{\text{Clay}}(u_1,u_2 | \theta) = (u_1^{-\theta} + u_2^{-\theta} - 1)^{-1/\theta}.
\end{equation}

The Clayton copula can be extended to include the range $\theta \in [-1,\infty)$, but for $\theta < 0$, $U_1$ and $U_2$ are negatively correlated, which is not a setting of interest for us. 
%
\subsection{Survival Copula and Survival Function}
\begin{figure}
\centering
\includegraphics[width=0.15\textwidth]{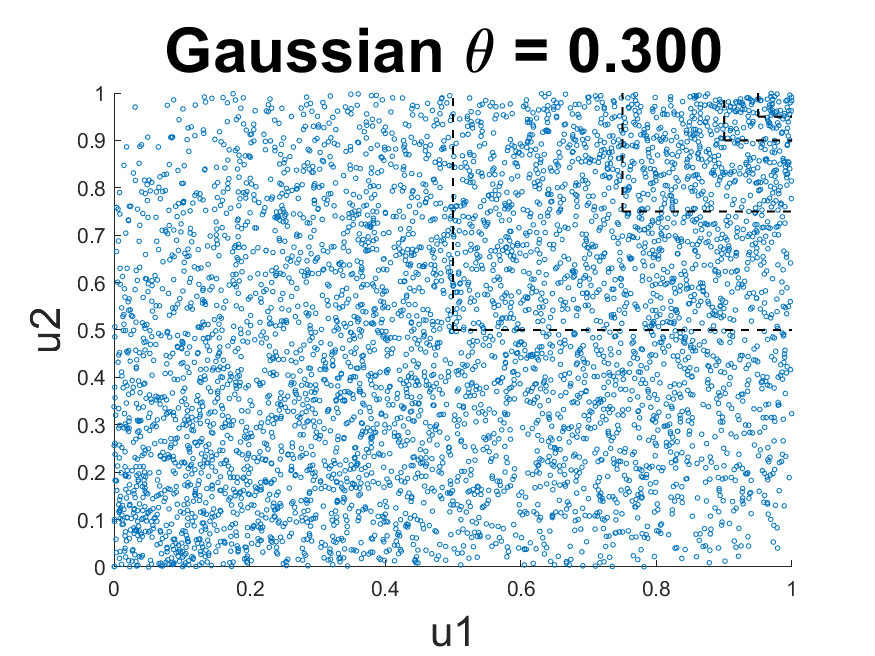}
\includegraphics[width=0.15\textwidth]{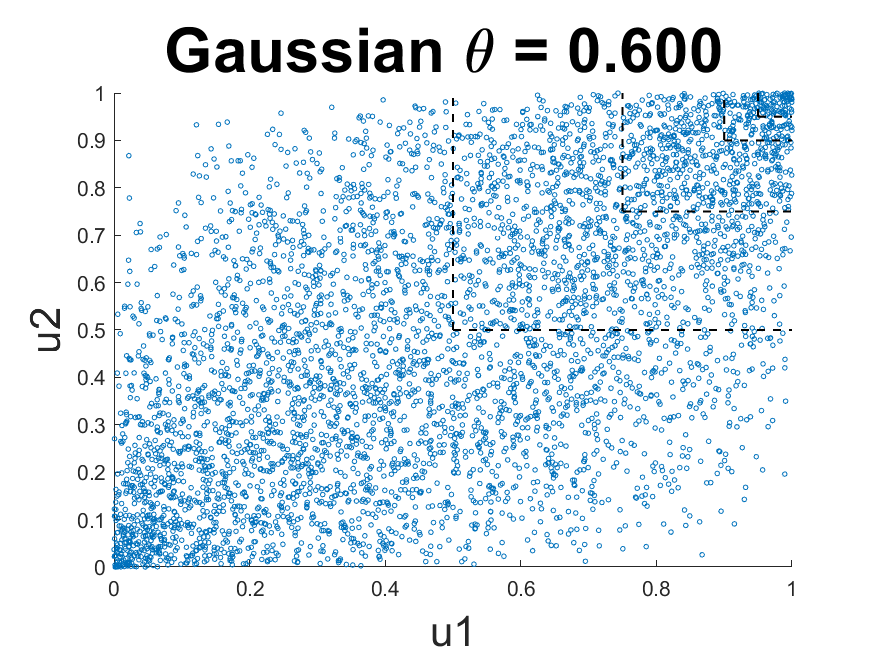}
\includegraphics[width=0.15\textwidth]{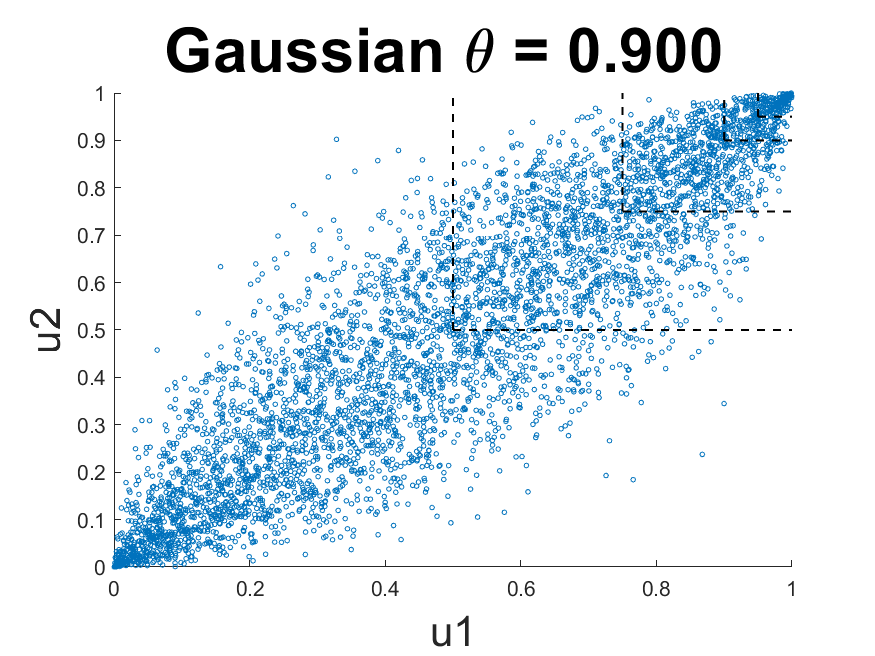}
\includegraphics[width=0.15\textwidth]{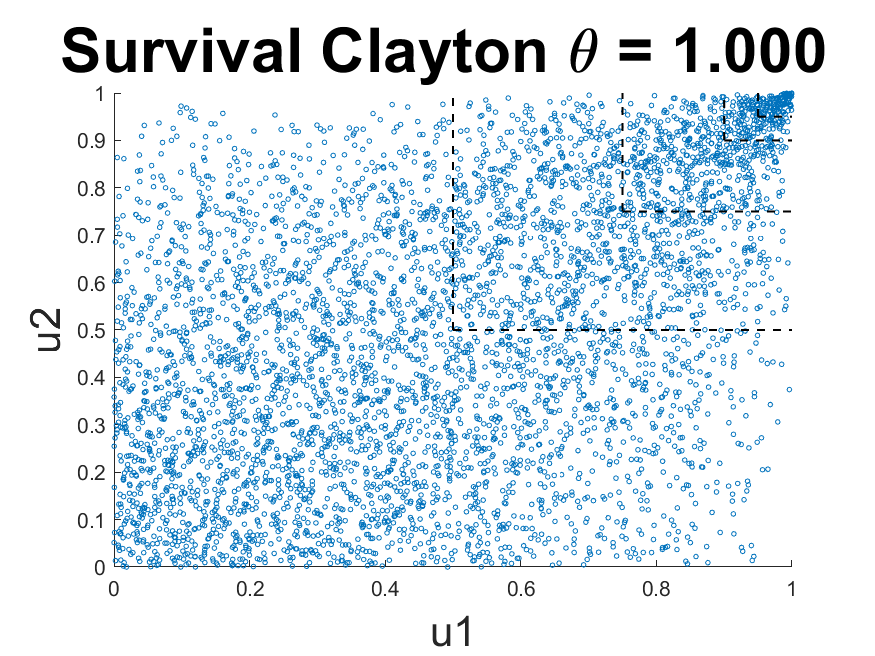}
\includegraphics[width=0.15\textwidth]{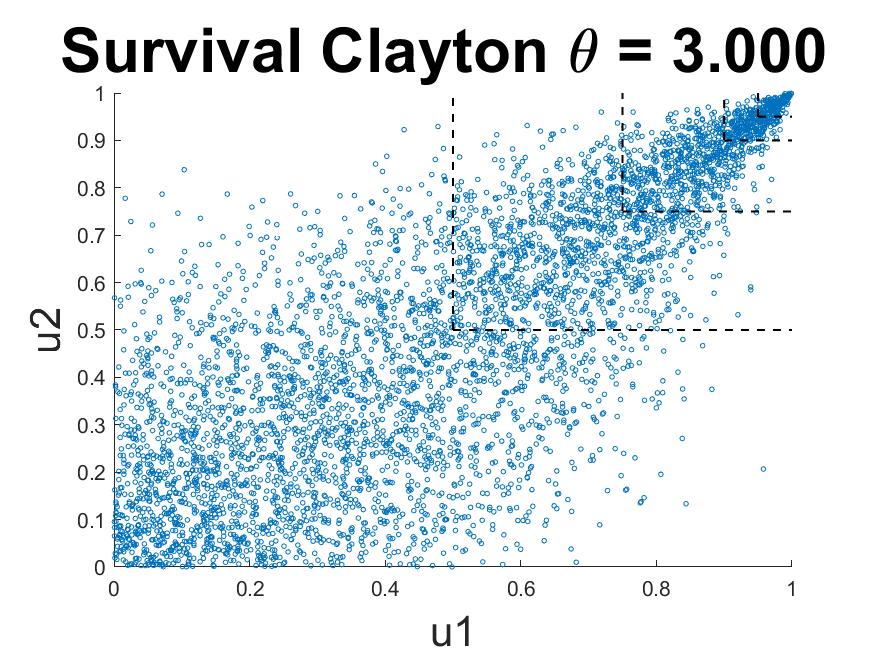}
\includegraphics[width=0.15\textwidth]{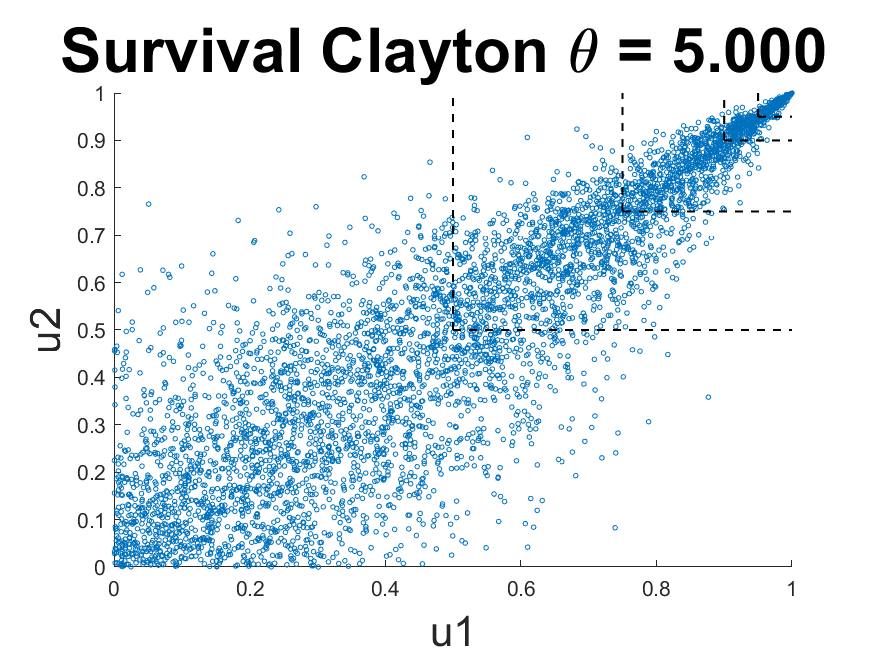}
\caption{\label{fig:generated} Example draws from the Gaussian and survival Clayton distributions of size $5000$. The dotted lines highlight the $q = 0.50, 0.75, 0.9$ and $0.95$ upper quantiles, subsets at which upper quadrant distributions will be defined. 
Note the survival Clayton copula has significantly more tail dependence than the Gaussian copula as shown by the tighter point concentrations in the upper quantiles.}
\end{figure}

The Clayton copula above 
has large dependence at the $(0,0)$ corner. We instead define the so called `survival Clayton' (SC) copula, in which the `survival' relation holds for any copula, as: 
%
\begin{equation}\label{eq:Survival}\CSC{}(u_1,u_2 | \theta) = C_{\text{Clay}}(1 - u_1, 1 - u_2 | \theta) + u_1 + u_2 - 1, \end{equation}
%
%
and with copula density $\cSC{}(u_1,u_2 | \theta) = c_\text{Clay}(1-u_1,1-u_2 | \theta)$.
%
Then, for $(U_1,U_2)$ distributed as $\CSC{}$ the   associated survival function (for the survival Clayton copula), $\SSC{}(u_1,u_2| \theta): = P_{SC}(u_1 > U_1, u_2 > U_2| \theta)$, is defined as:
\begin{equation}\label{eq:SurvivalFunc}
\begin{split}
 \SSC{}(u_1,u_2| \theta) &=  C_\text{Clay}(1 - u_1, 1 - u_2| \theta) \\
 &=\CSC{}(u_1,u_2| \theta) + 1 - (u_1 + u_2)
\end{split}
\end{equation}
%
See figure \ref{fig:generated} for sample draws from both the Gaussian copula and the survival Clayton copula.
%

%
%
%
%
\section{Methodology}
A motivating factor in analyzing anomaly scores with copulas is the removal of confounding marginal structure, especially in the tail where the precise marginal distribution is most difficult to estimate. We are interested in the similarity between variables, especially in the tails, not the marginal score distributions. And for anomaly scores in particular, we are only concerned with the score relationship in the upper quadrant only. 
Although where to define the upper quadrant requires further consideration below, working with copulas  simplifies the situation through marginal standardization. 

Let $R_q$ denote the (upper) q-quadrant $R_q=[q,1]^2. $ 
As we are primarily interested in associations within $R_q$ we might attempt to fit the survival Clayton copula density $c_{SC}(\cdot|\theta)$ only to sample points that lie in $R_q$ after marginal standardization, and estimate the association parameter $\theta$. 
However, this has the undesirable property of simply favoring copula density models that have the greatest density inside of $R_q$, regardless of the actual shape or fit within $R_q$.

%
%
%

  We thus propose deriving the conditional density specifically for the upper copula quadrant, and using properties of that density to define a similarity measure.
Specifically, we define the q-quadrant conditional density of the survival Clayton copula as 
\begin{equation}\label{eq:qc}
cq_{SC}(u|q,\theta) = \frac{\cSC{}(u_{1},u_{2}| \theta)}{\SSC{}(q,q|\theta)} I\{(u_{1},u_{2})\in R_q\}.
\end{equation}
%
We call the function $cq_{SC}(\cdot)$ the survival Clayton conditional (quantile or quadrant) density, and we note that it is not itself a copula, it is a conditional density derived from a copula. In particular the marginal distributions of $cq_{SC}(\cdot)$ are not uniform.


In practice, we do not expect samples directly from a copula. Instead, for a random sample $\mathbf{Y} = \{ (y_{i1}, y_{i2}) \}_{i=1}^n$, we apply variable-wise transformations \citep{ruppert2011statistics} and define a semiparametric pseudo log-likelihood function as
\begin{equation}\label{eq:psuedo}
\ell(\theta|\mathbf{Y},q) = 
\sum_{i \in R_q} \text{log} \,[ cq_{SC}(\hat u_{i1}, \hat u_{i2}|q,\theta)],
\end{equation}
in which 
\begin{equation}\label{eq:empCDF}\hat{u}_{ij} = \hat{F}_{Y_j}(y_{ij}) = 
 \frac{1}{n+1} \sum_{k=1}^n I \lbrace y_{kj} \leq y_{ij}  \rbrace.\end{equation}
 
 and $i \in R_q$ is shorthand for $\lbrace i |\hat{\mathbf{u}}_i \in R_q \rbrace$.

\subsection{Copula Quadrant Similarity} 
\begin{figure}
\centering
\includegraphics[width=\hcolw]{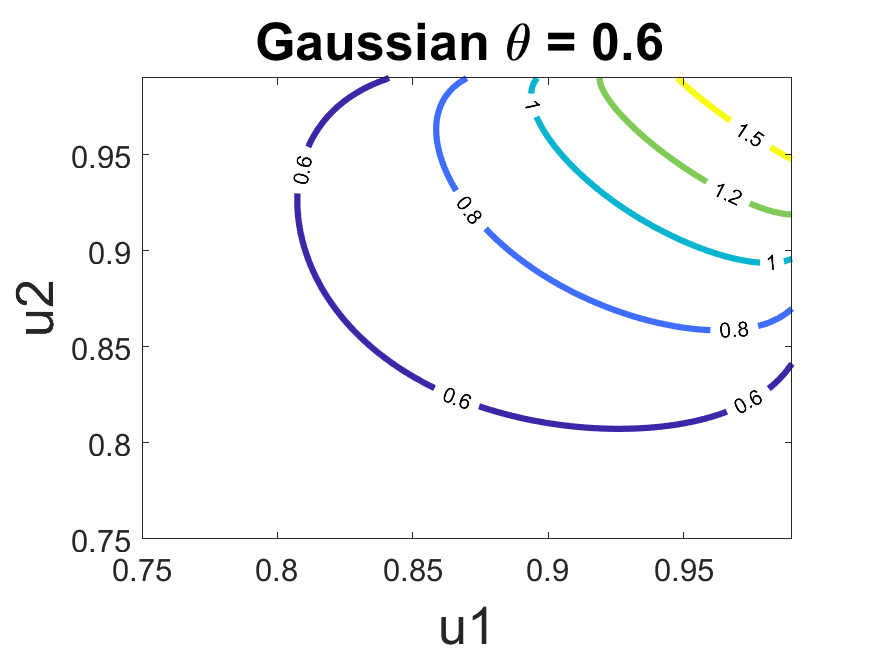}
\includegraphics[width=\hcolw]{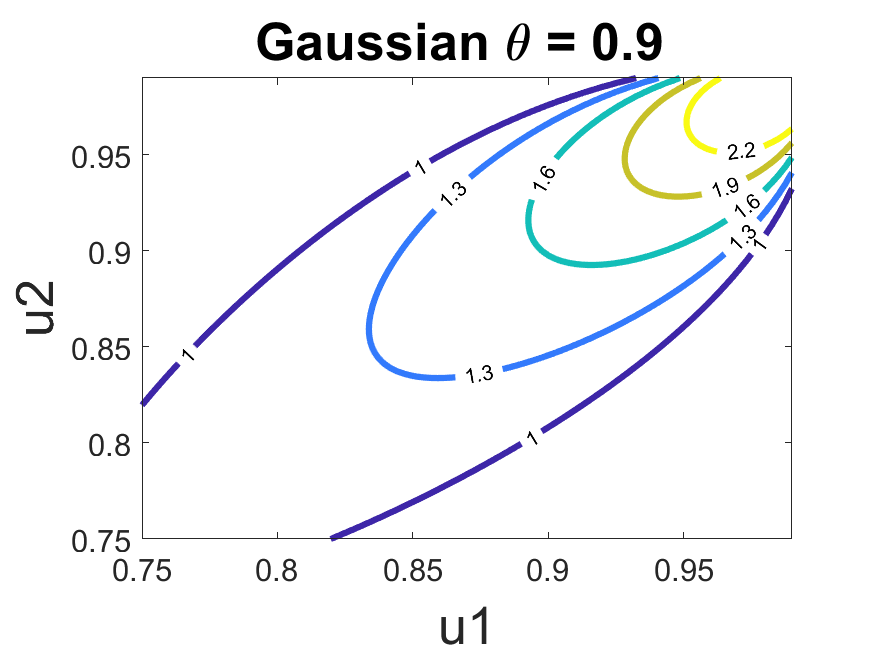}
\includegraphics[width=\hcolw]{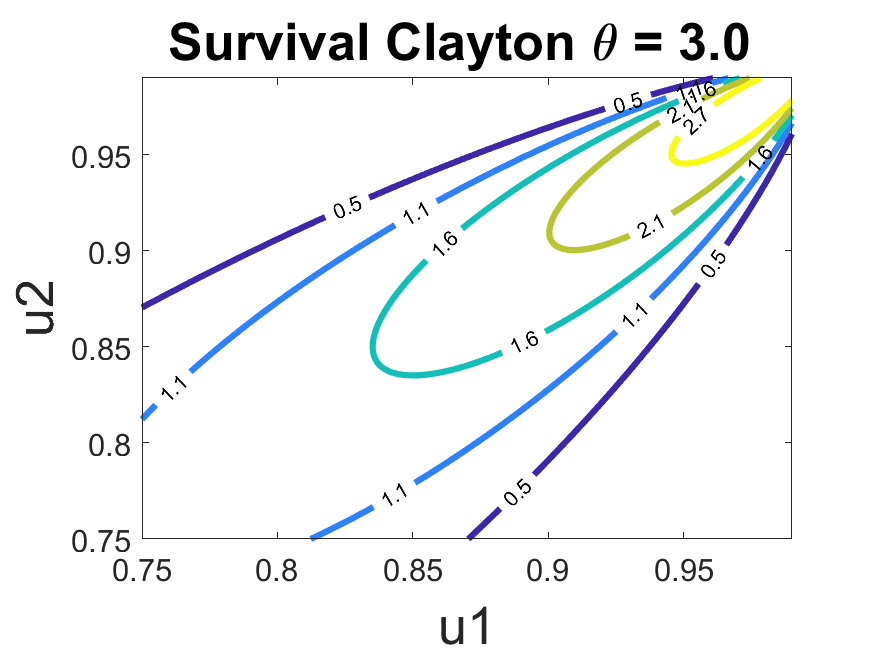}
\includegraphics[width=\hcolw]{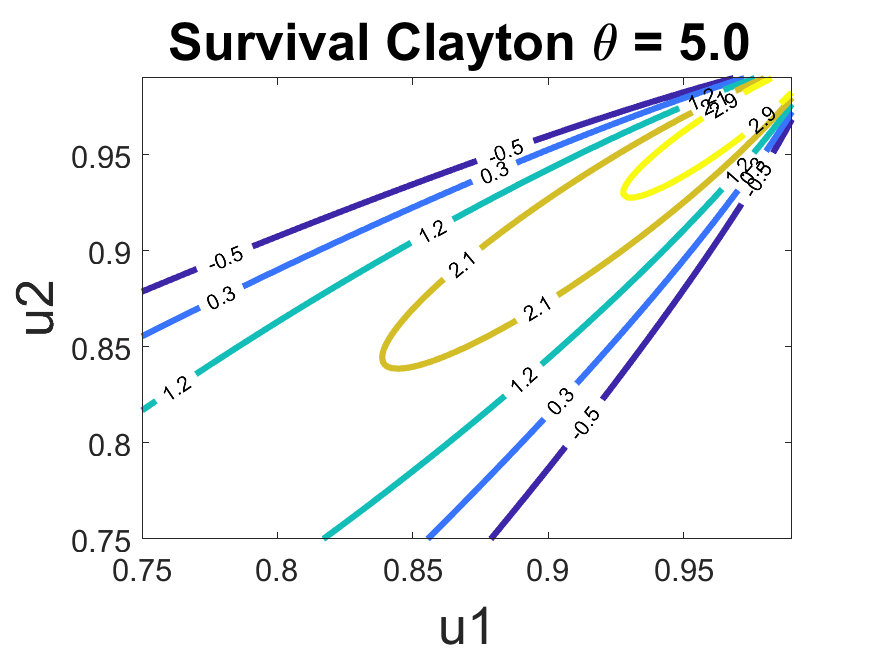}
\caption{\label{fig:Contours} Contours of the log-densities for two Gaussian and two survival Clayton Copulas. The survival Clayton contours vary much more significantly than the Gaussian, illustrating the fact that they are much more sensitive to tail dependence.}
\end{figure}

\begin{figure}
\centering
\includegraphics[width=\hcolw]{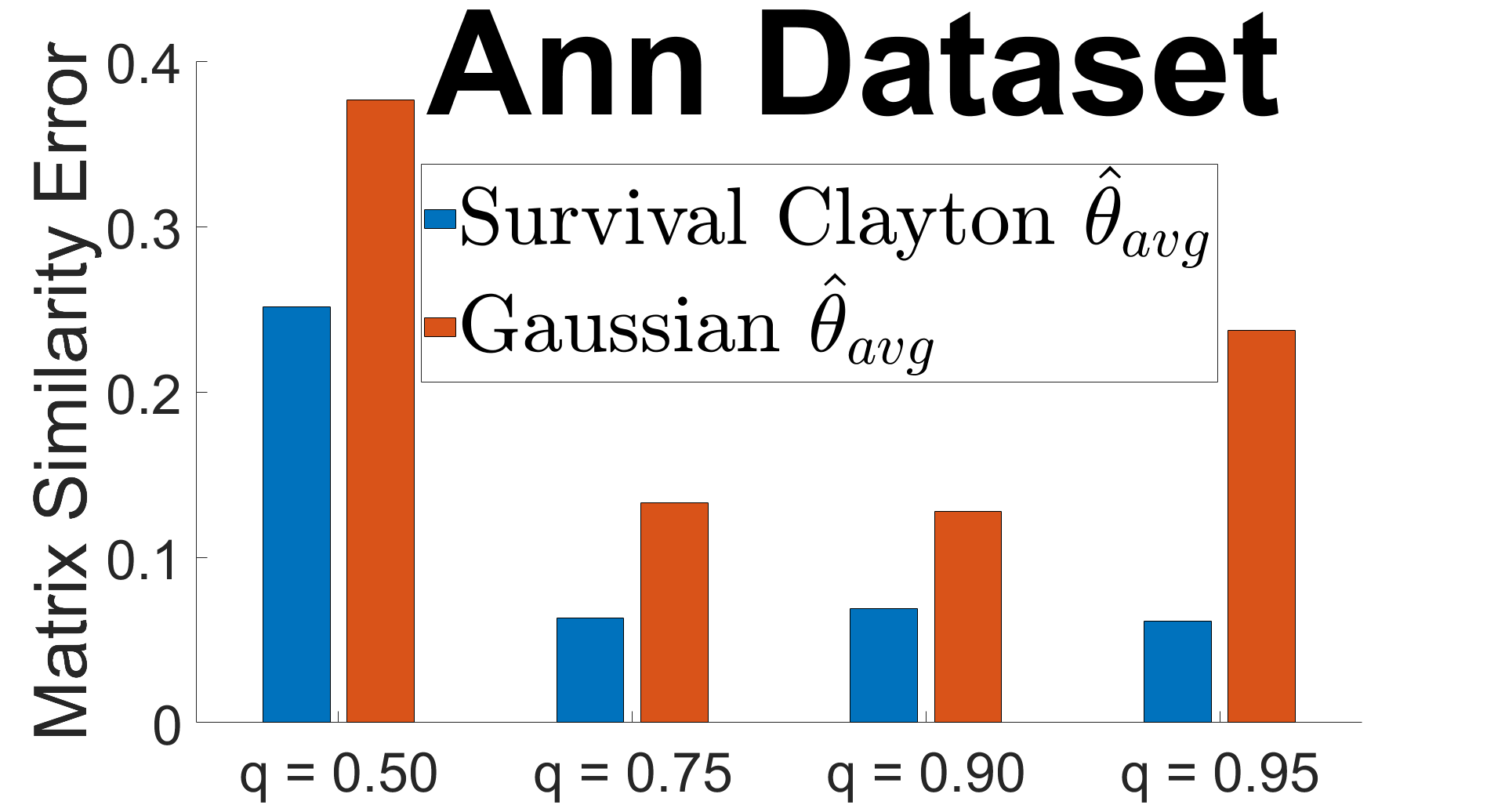}
\includegraphics[width=\hcolw]{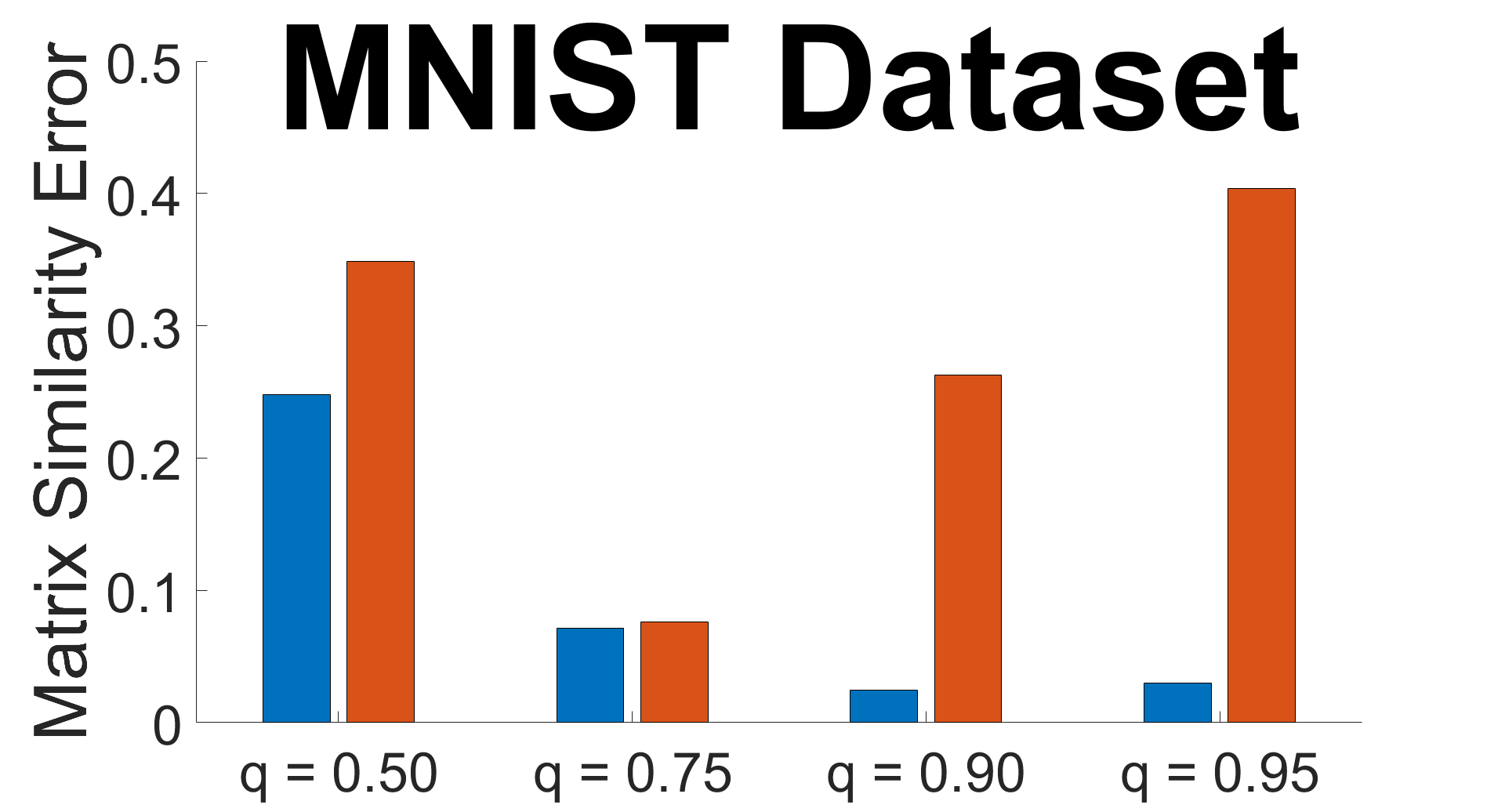}
\includegraphics[width=\hcolw]{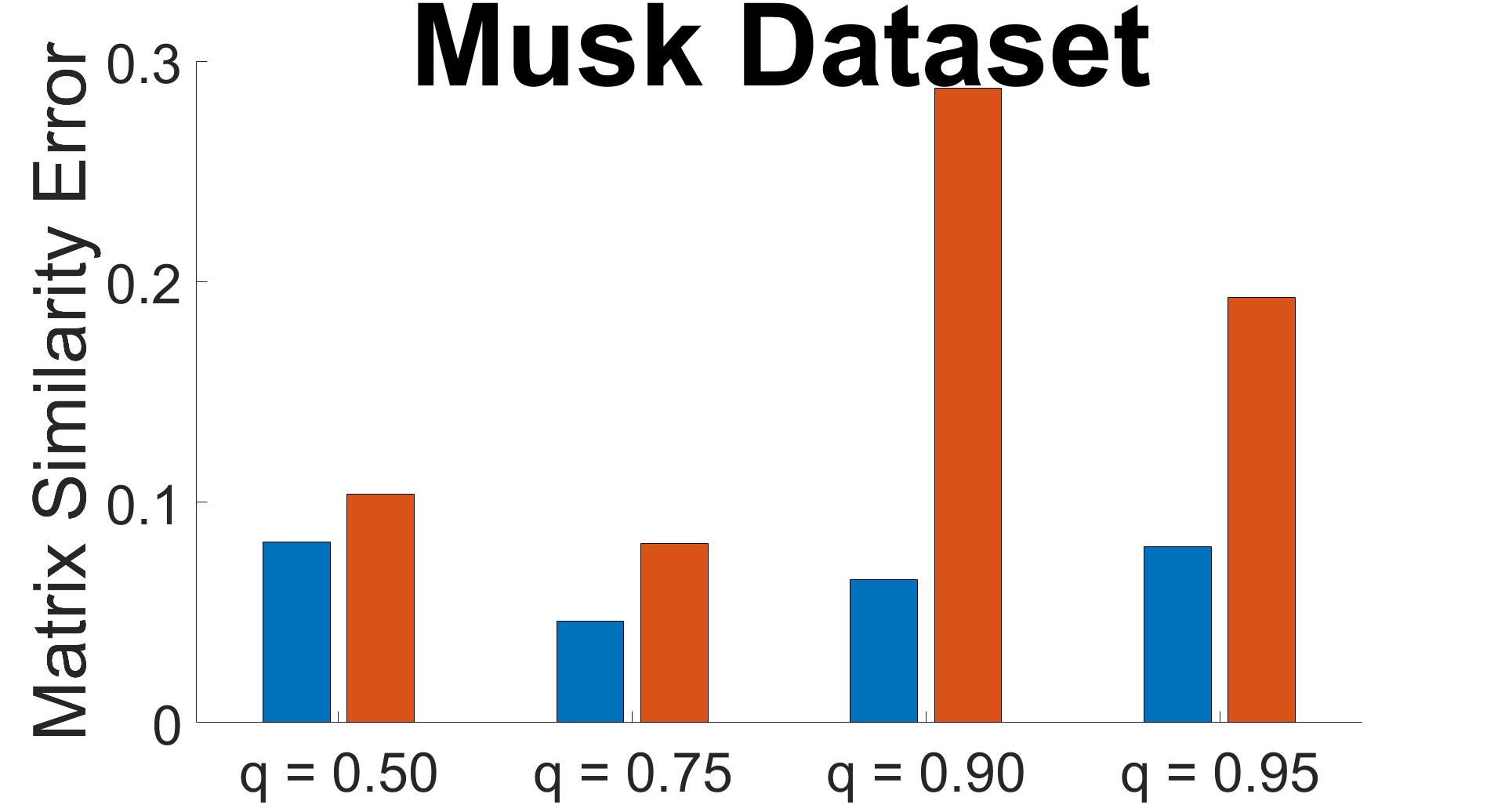}
\includegraphics[width=\hcolw]{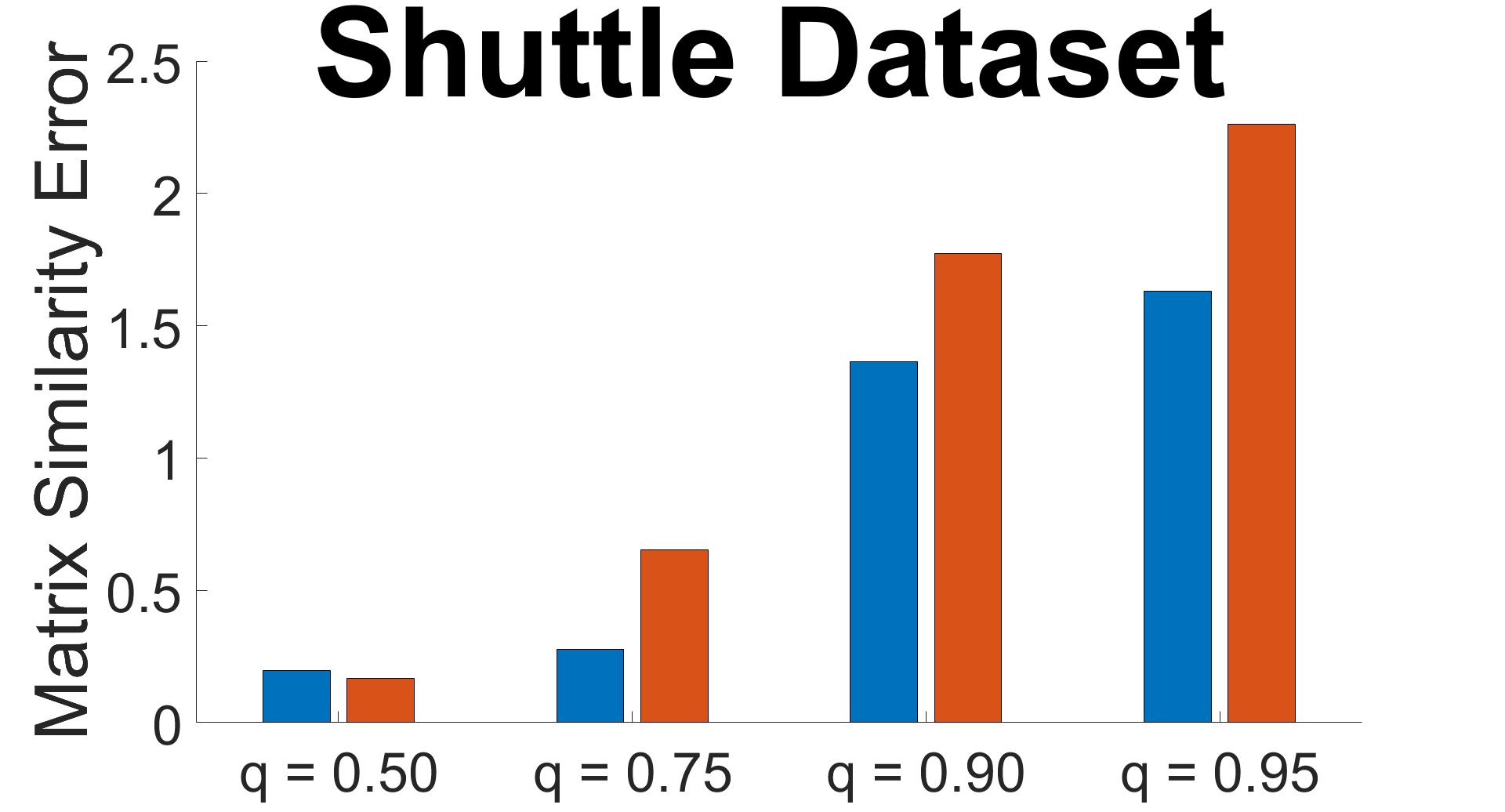}
\caption{\label{fig:ClayGauss} These plots show the dominance of the survival Clayton Copula over the Gaussian Copula. This is due to tail dependence nature of the survival Clayton copula, it is sensitive to small changes in the upper tail region, which is ideal for associating the important anomalies/extremes.}
\end{figure}

In order to create a similarity measure that is especially sensitive to upper tail dependence and agnostic to lower quadrant dependence, we utilize  flexible copulae that can capture increasingly strong tail dependence. The survival Clayton copula exhibits varying degrees of increasing tail dependence as its single association parameter $\theta$ increases, and is shown to have superior tail dependence sensitivity to the Gaussian copula, see figures \ref{fig:Contours} and \ref{fig:ClayGauss}. In figure \ref{fig:Contours} we see the Clayton copula has faster changes of contours (both larger directional derivatives, as well as just generally higher density in this upper quadrant). For instance consider the directional derivative of the likelihoods in the [1,-1] direction: the Gaussian contours are nearly parallel to this, but the Clayton contours vary considerably, which shows they are sensitive to how dependent data are, that is, how close the data are to x=y line. Additionally, the Clayton copula possesses beneficial theoretical properties for modelling tail dependence: it is a natural limit for conditional bivariate extremes, which also have an Archimedean copula dependence structure; and is the only copula which is invariant under the upper q-quadrant conditioning we described above  \citep{juri2002copula, oakes2005preservation, charpentier2003tail}. 
%

To measure extremal similarity we propose using the association parameter $\theta$ that maximizes the  log-likelihood given in equation \ref{eq:psuedo}, for a specific $q$: 
\begin{equation}\label{eq:thetaS}\ts{}(q) = \text{argmax}_{\theta \in \mathbb{R}^+} \lbrace \ell(\theta|\mathbf{Y},q) \rbrace. \end{equation}
%
Although we have chosen the survival Clayton copula as our base measure, our definitions allow for the flexibility of choosing among other copula families, as long as an association parameter $\theta$ is indicative of tail dependence. The proposed $\ts{}$ will capture information about the shape of the tail dependence, how quickly it strengthens inside of $R_q$. We find below that $\ts{}$ does well empirically for most values of $q$, thus we propose using it as the first component of our similarity measure.

A weakness of $\ts{}$ is that it does not explicitly 
account for the fraction $n_q/n$ of points in $R_q$, where $n_q = \# \{\hat{\mathbf{u}}_i \in R_q\}$.
Whereas the empirical copula $\hat{C}$ and empirical survival function $\hat{S}$, which are used to define  $\bar{\chi}$, do \citep{coles1999dependence}. They are defined as
 \begin{equation}
 \begin{split}
\label{eq:empCop} \hat{C}(q_1,q_2) & = \frac{1}{n} \sum_{i=1}^n I \lbrace \hat{u}_{i1} \leq q_1 , \hat{u}_{i2} \leq q_2 \rbrace \\
\hat{S}(q_1,q_2) & = \frac{1}{n} \sum_{i=1}^n I \lbrace \hat{u}_{i1} > q_1,\hat{u}_{i2} > q_2 \rbrace, 
 \end{split}
\end{equation}
%
with $\hat{S}(q,q) = n_q/n$,
and $\hat{S}(q_1,q_2) \approx \hat{C}(q_1,q_2) + 1 - (q_1 + q_2)$.
This last relation is only approximate due to both the finite and random nature of a sample. When using $\hat{u}_{ij}$ derived from the empirical distribution function (see equation \ref{eq:empCDF}), then there is at most an error of $1/n$. 

As an alternative copula quadrant similarity measure we propose matching the survival Clayton survival function with the empirical survival function to find the best association parameter $\theta$.
We define 
%
\begin{equation}\label{eq:thetaF}\tf{}(q) = \text{argmin}_{\theta \in \mathbb{R}^+}  \lbrace |\SSC{}(q,q| \theta) - \hat{S}(q,q)|  \rbrace. \end{equation}
%
There are many copulas that exhibit increasing tail dependence through increasing their sole association parameter $\theta$ (e.g., Joe, Clayton, Gumbel, Gaussian), with independence at a minimum and perfect dependence at a maximum (possibly infinite). For such copulae, the function $S(q,q|\theta)$ is strictly monotonically increasing in $\theta$. Thus as long as $\hat{S}(q,q) > (1-q)^2$, that is there are more q-quadrant points than expected under independence, then equation \ref{eq:thetaF} can be solved as an equality instead of minimization:
\begin{equation}\label{eq:thetaF2} \tf{}(q) = \lbrace \theta | \SSC{}(q,q| \theta) = \hat{S}(q,q) \rbrace. \end{equation}
However, when $\hat{S}(q,q) < (1-q)^2$, then the solution to equation \ref{eq:thetaF}, cannot be solved as an equality, but instead will have a boundary $\theta$ as the minimizer, which is where the copula family converges to the independent copula for the above mentioned copula families. In addition if $\hat{S}(q,q) > 1- q$ then equation \ref{eq:thetaF} cannot be solved as an equality, and in fact has no solution if the domain of $\theta$ is unbounded (informally $\theta = \infty$ is the solution). However it is guaranteed that $\hat{S}(q,q) < 1- q$ if $\hat{\mathbf{u}}$ is computed using equation $\eqref{eq:empCDF}$. One option if the $\hat{\mathbf{u}}$ is not computed using $\eqref{eq:empCDF}$ is to bound the domain of $\theta$, at least for the minimization in equation \ref{eq:thetaF}.

The similarity measures $\tf{}(q)$, $\hat\chi(q)$, and $\hat{\bar{\chi}}(q)$ are all based on a single measurement at $(q,q)$ of the empirical survival function or empirical copula. The proposed $\tf{}(q)$ is model based, allowing it to inherit the tail dependent properties of the chosen model (i.e., the survival Clayton model), whereas $\chi$ and $\bar{\chi}$ are model-free estimates of similarity using only $\hat{C}(q,q)$ and $\hat{S}(q,q)$. Model based estimates such as $\tf{}$ and $\ts{}$ tend to have less variation than model free ones, and exhibit the tail dependence sensitivities of the base copula model from which they are derived from.

These two proposed estimators $\ts{}$ and $\tf{}$ complement each other well, capturing information about the  strength of tail dependence inside of $R_q$ and the excess fraction inside of $R_q$, respectively. The first, $\ts{}$, is able to take into account tail shape (and dependence strength) information, but breaks down at larger $q$, where there is not enough data to reliably estimate tail dependence. For these larger quantiles $\tf{}$ will be a more reliable estimator. In addition, assuming the survival Clayton model is well specified, the two estimators are consistent and asymptotically independent, which is proved in the Appendix \citep{khan2004approximation}. Noting this asymptotic independence and the complementary strengths of both estimators, we define our \emph{Copula Quadrant Similarity Measure} as the average of these two measures: 
\begin{equation} \label{measure}  \ta{}(q) := \frac{1}{2}\ts{}(q) + \frac{1}{2}\tf{}(q). \end{equation}
Larger $\ta{}$ values correspond to higher tail dependence or greater anomaly or extremal similarity, whereas value near $0$ imply tail independence.

\begin{figure}
\centering
\includegraphics[width=\hcolw]{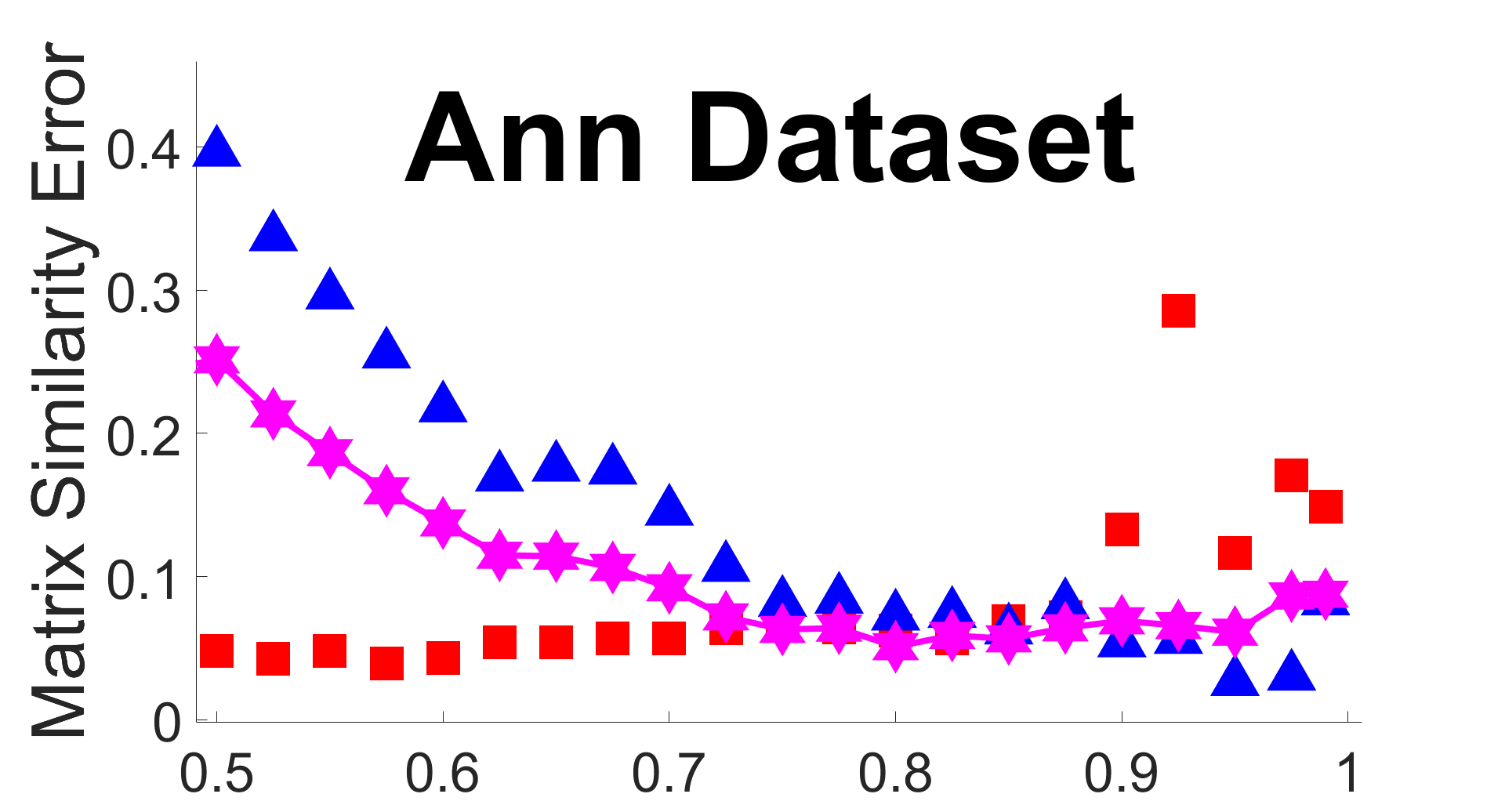}
\includegraphics[width=\hcolw]{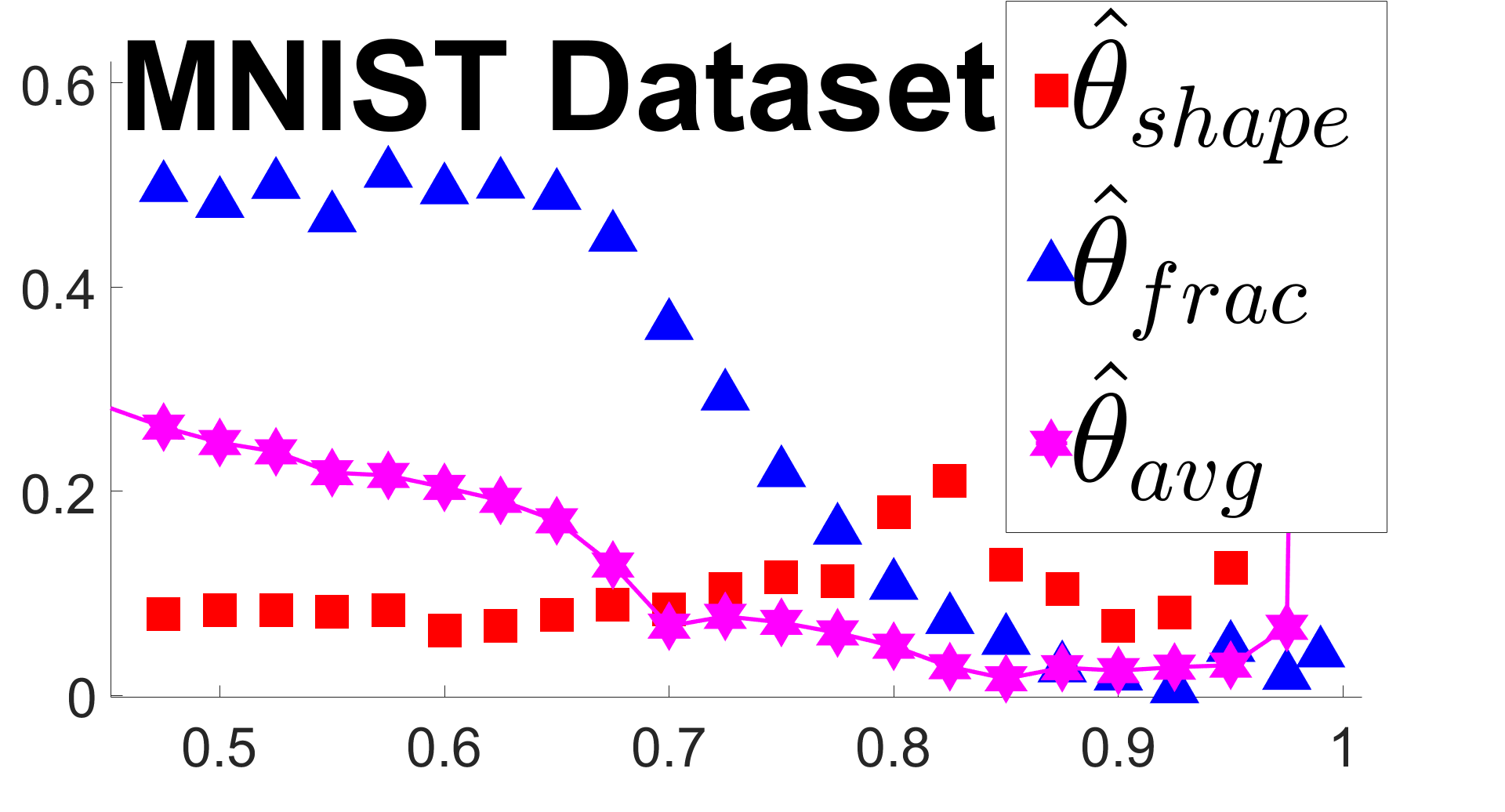}
\includegraphics[width=\hcolw]{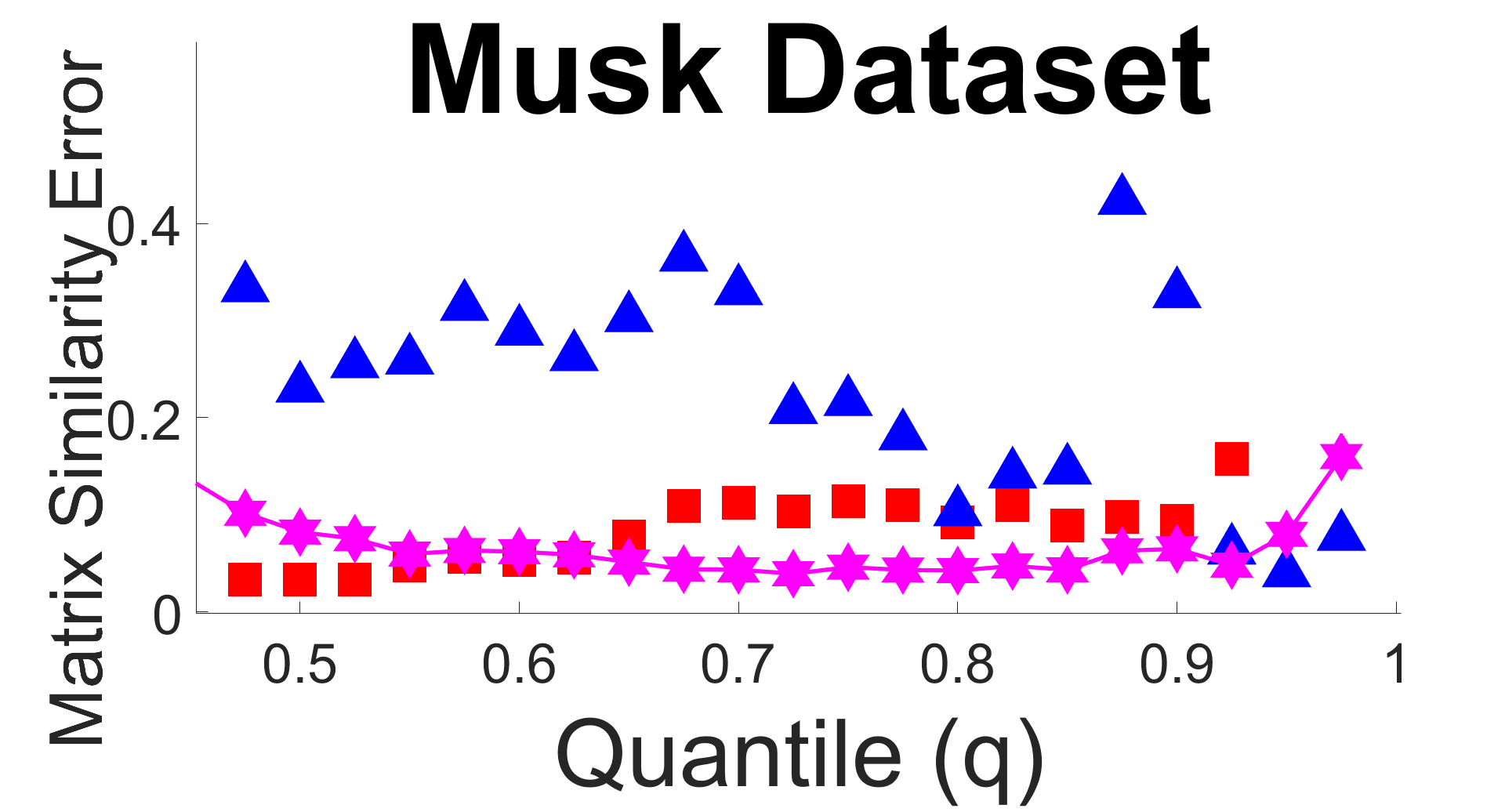}
\includegraphics[width=\hcolw]{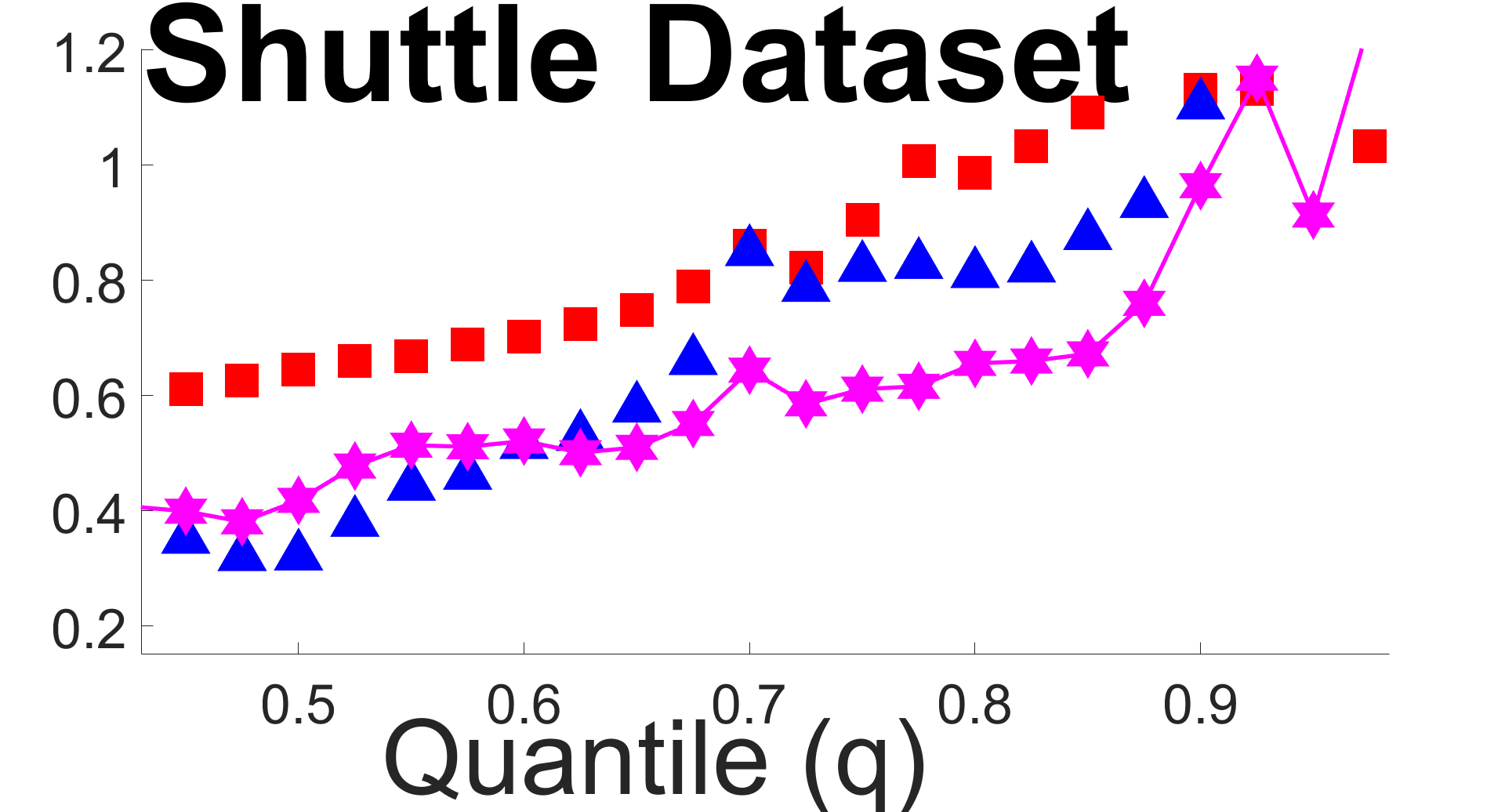}
\caption{\label{fig:qZoom}  These plots show the complementary nature between $\tf{}$ (blue) and $\ts{}$(red), and the superiority of their average, $\ta{}$ (purple) across four datasets. We see that $\ta{}$ performs well over a broad range of $q$, and this average even outperforms each component individually at some $q$ values, such around $q = 0.8$ for these datasets. The $\ts{}$ measure performs well for most values of $q$, but performs poorly as $q$ approaches $1$. However $\tf{}$ has the opposite behavior, it's performance improves for larger values of $q$. For this reason it is clear that a weighted average dependent on $q$ would be optimal (at lower $q$, $\ts{}$ should be emphasized, at higher $q$, $\tf{}$ is superior), however for simplicity equal weights have been used for all $q$.}
\end{figure}

The advantage of averaging the two estimators as opposed to either one individually is shown in figure \ref{fig:qZoom}. We see that the average is more stable over a broad range of $q$, and often outperforms either one individually for $q$ values in desired range around $q = 0.75$.
Empirical results shown below further demonstrate the effectiveness of $\ta{}$ as a measure of extremal or anomaly similarity, capturing the essential tail dependence through the survival Clayton copula. 
%
%
\section{Evaluation Details}
A visual summary of the proposed methodological flow is shown in figure \ref{fig:copFlow}. The rest of this section goes over the details of each step.
%

%
%
\subsection{Anomaly Score Similarity}
We describe how this procedure is applied to measure the similarity between pairs of anomaly detection scoring algorithms. Anomalies are present in a given dataset $D := \lbrace \mathbf{x}_i \rbrace_{i=1}^n$ where each $\mathbf{x}_i$ is a datapoint in some not necessarily continuous space $\mathcal{X}$. We have access to $k$ anomaly detection algorithms $\mathcal{A}_1 \cdots \mathcal{A}_k$, such that $\mathcal{A}_j(x_i)$ is the anomaly score given by algorithm $j$ on point $\mathbf{x}_i$ (the higher the score, the more anomalous). We denote the matrix $\mathbf{Y}: y_{ij} := \mathcal{A}_j(\mathbf{x}_i)$. Then the procedure discussed below can be applied, where the similarity between the pair of vector scores $\mathbf{y_i},\mathbf{y_j}$ represents the similarity of algorithms $\mathcal{A}_i$ and $\mathcal{A}_j$.
%

Using this dataset of scores,  $\mathbf{Y}$, we compute the pairwise column similarity. The first step is to transform the data $\mathbf{Y}$ onto the copula scale $\mathbf{U}$ using the empirical cdf, equation \ref{eq:empCDF}. A $q$ value is chosen, we make the general recommendation of $ q = 0.75$, but this can vary depending on application. For each pair of column indexes ${i,j}$, the conditioned copula MLE estimator $\ts{}(q)$ is fit on the pair [$\hat{\mathbf{u}}_i,\hat{\mathbf{u}}_j$] using any convex optimization method (as the Clatyon copula density is convex in $\theta$). Similarly the estimator $\tf{}(q)$ is fit on each pair by solving equation \ref{eq:thetaF2}. The similarity measure is defined as in equation \ref{measure}, as the average of the two estimators. We denote $W$ to be the matrix of pairwise similarity measures, $w_{ij}$ is the similarity measure $\tf{}(q)$ applied to columns $\mathbf{y_i}$ and $\mathbf{y_j}$.
\subsection{Alternative Similarity Measures}
We want to compare our similarity measure, $\ta{}(q)$, with three main alternatives, $\chi(q)$, $\bar{\chi}(q)$ and ``upper correlation" or ``UCorr".  $\chi(q)$ and $\bar{\chi}(q)$ were described in the introduction, see equations \ref{eq:Chi} and \ref{eq:ChiBar}. We apply these to data by taking the empirical versions; the probabilities in equations \ref{eq:Chi} and \ref{eq:ChiBar} are replaced with the fraction of data in the relevant regions.  ``UCorr" is the correlation of the data in $R_q$, which is called ``uppertail conditional rank correlation” by \cite{charpentier2003tail}. These alternatives will have the same flow as $\ta{}(q)$, in particular they are applied to the transformed variables $\hat{\mathbf{U}}$ and fit pairwise.
%
%
\begin{figure}
\centering
\includegraphics[width=0.5\textwidth]{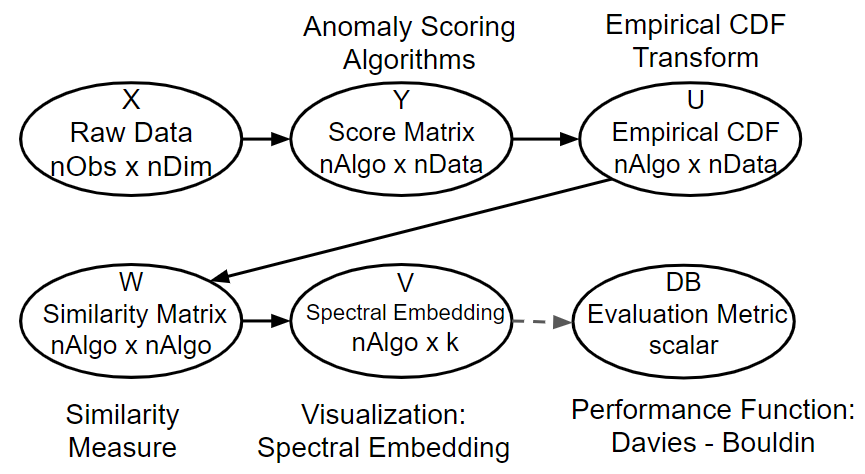}
\caption{\label{fig:copFlow}  Flow diagram of the methodology, starting from an anomaly detection dataset and ending with visualizations and similarity measure performance metrics. The dotted line emphasizes that the performance measure can only performed if there is a ground truth clustering expected.}
\end{figure}
\subsection{Spectral Clustering Visualization and Evaluation Metric}
There is no clear choice to decide how good a certain similarity measure is since in general this is an unsupervised task. The approach we will take is to assume there is some ground truth clustering present. In the anomaly detection algorithm case this corresponds to the belief that a certain set of anomaly detection algorithms are similar to each other in that they give extreme scores to the same set of observations. In the data matrix case the belief is that certain variables (columns) are related to each other by having strong tail dependence. In order to evaluate our similarity measure using these assumed cluster assignments, we will first use our similarity measure to build a spectral embedding. A good description of spectral embedding is given by \cite{von2007tutorial}. 
%
%
Using the spectral embedding constructed from the pairwise similarity matrix $W$, we can now quantitatively measure the performance of similarity measures when there is a clustering that is taken as ground truth. While there are many clustering objectives/loss functions to choose from, we employ the Davies-Bouldin index. The Davies-Bouldin index roughly measures the proportion of inter-cluster spread versus intra-cluster spread, see the Appendix. 
\section{Simulations and Experiments}
\subsection{Block Dataset}
\begin{figure}
\centering
\includegraphics[width=0.21\textwidth]{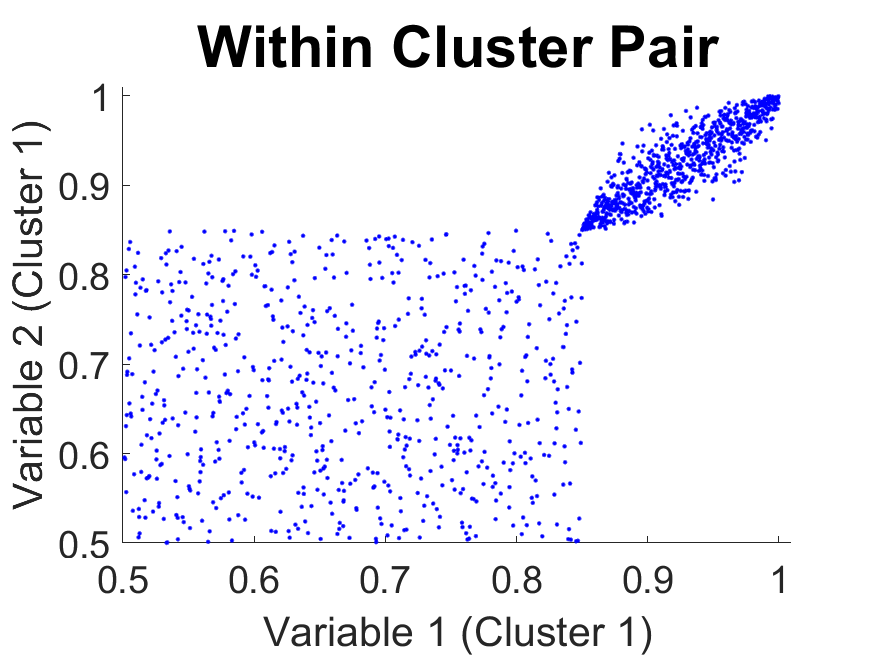}
\includegraphics[width=0.21\textwidth]{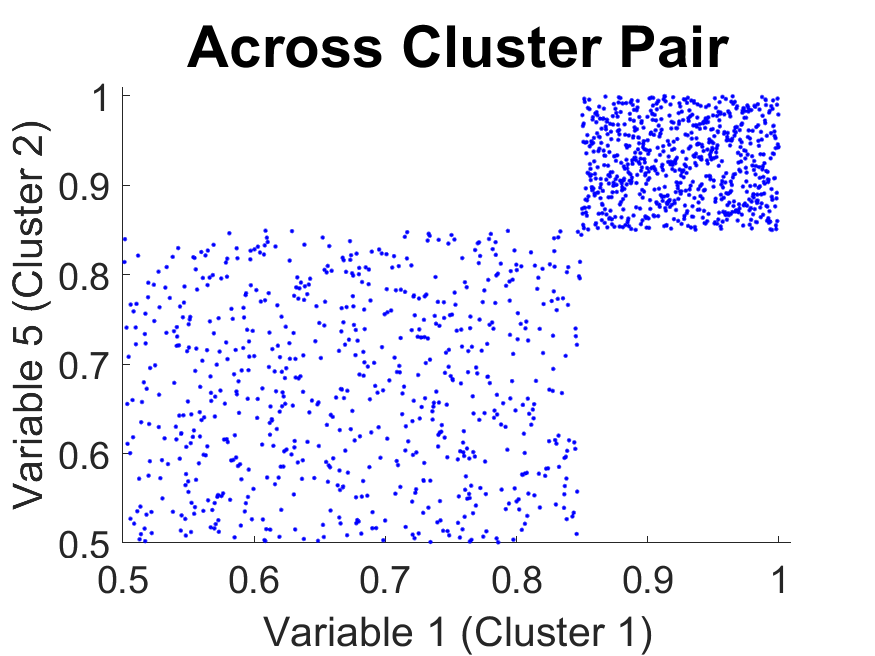}
\caption{\label{fig:BlockData} Two example realized pairs for the block model. The plots are zoomed in to emphasize the difference in the upper block, the lower blocks are identical for the two of them, uniform extending to the axes. (Left) Two columns both taken from the same cluster form a tightly dependent upper block. (Right) A pair of columns from different clusters have an independent upper block.}
\end{figure}

A scenario that is ideal to show the necessity of considering a similarity measure more complex than $\ch{}(q)$ or $\cbh{}(q)$ is data generated from two blocks in the unit square. The lower left block will be on $[0,b] \times [0,b]$ for $b = 0.85$, and data is uniformly distributed conditional on being inside the block. A more complex distribution is used for the upper block; we choose to create two clusters of four variables (columns) each. If a pair of variables come from the same cluster, then the upper block has the dependence of a Gaussian copula, see the left plot of figure \ref{fig:BlockData}. If the pair comes from variables of different clusters, than the two have an independent upper block, as in the right plot of figure \ref{fig:BlockData}.
For $q < b$, both $\ch{}(q)$ and $\cbh{}(q)$ will be insensitive to that choice of upper right block, they will assign the same similarity measure if the upper block has tight dependence or is independent because they are simply a function of counting the number of points inside that block. By contrast our $\ta{}(q)$ will strongly differentiate those two cases for any value of $q$, since the shape and strength of tail dependence in the upper right quadrant strongly affects the copula fit. Spectral embeddings using the aforementioned similarity measures are shown in figure \ref{fig:BlockData}, and the DB-indexes of these embeddings shown in table \ref{table:Main Results}. Both $\ch{}(q)$ and $\cbh{}(q)$ fail to differentiate the red and blue clusters, they do not assign different similarities to the left and right plots in figure \ref{fig:BlockData}.

%
\begin{figure}
\centering
\includegraphics[width=0.22\textwidth]{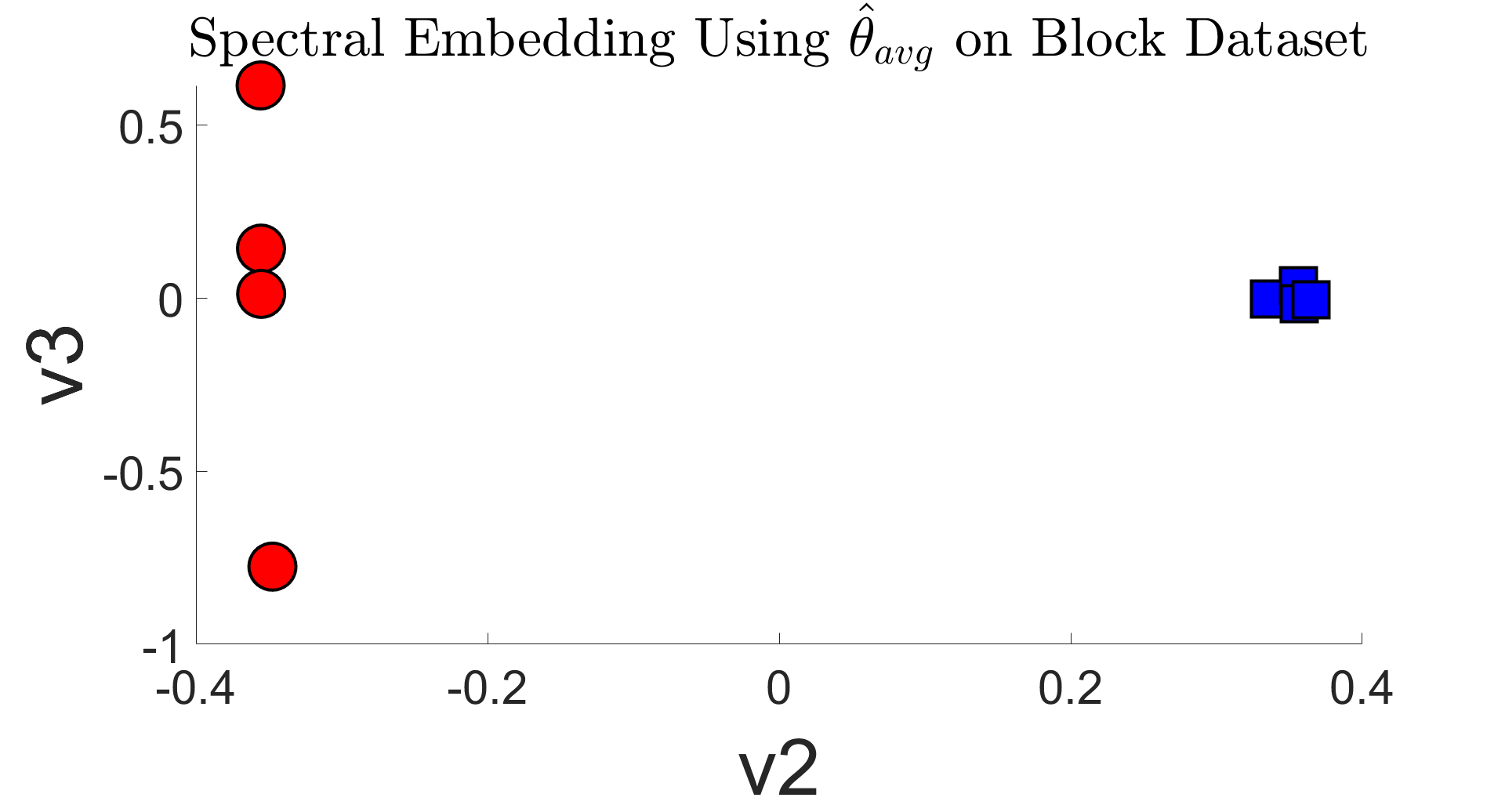}
\includegraphics[width=0.22\textwidth]{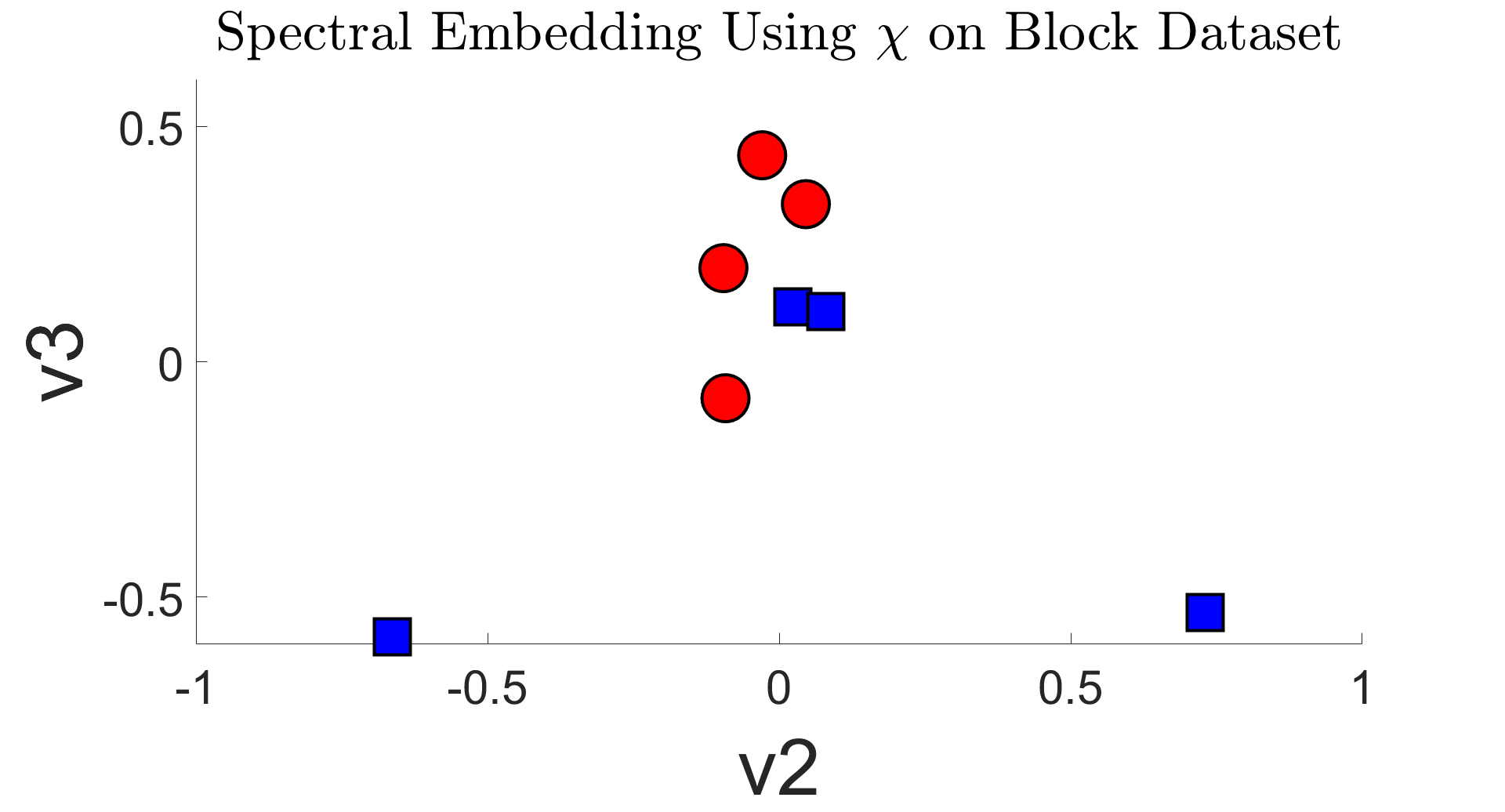}
\includegraphics[width=0.22\textwidth]{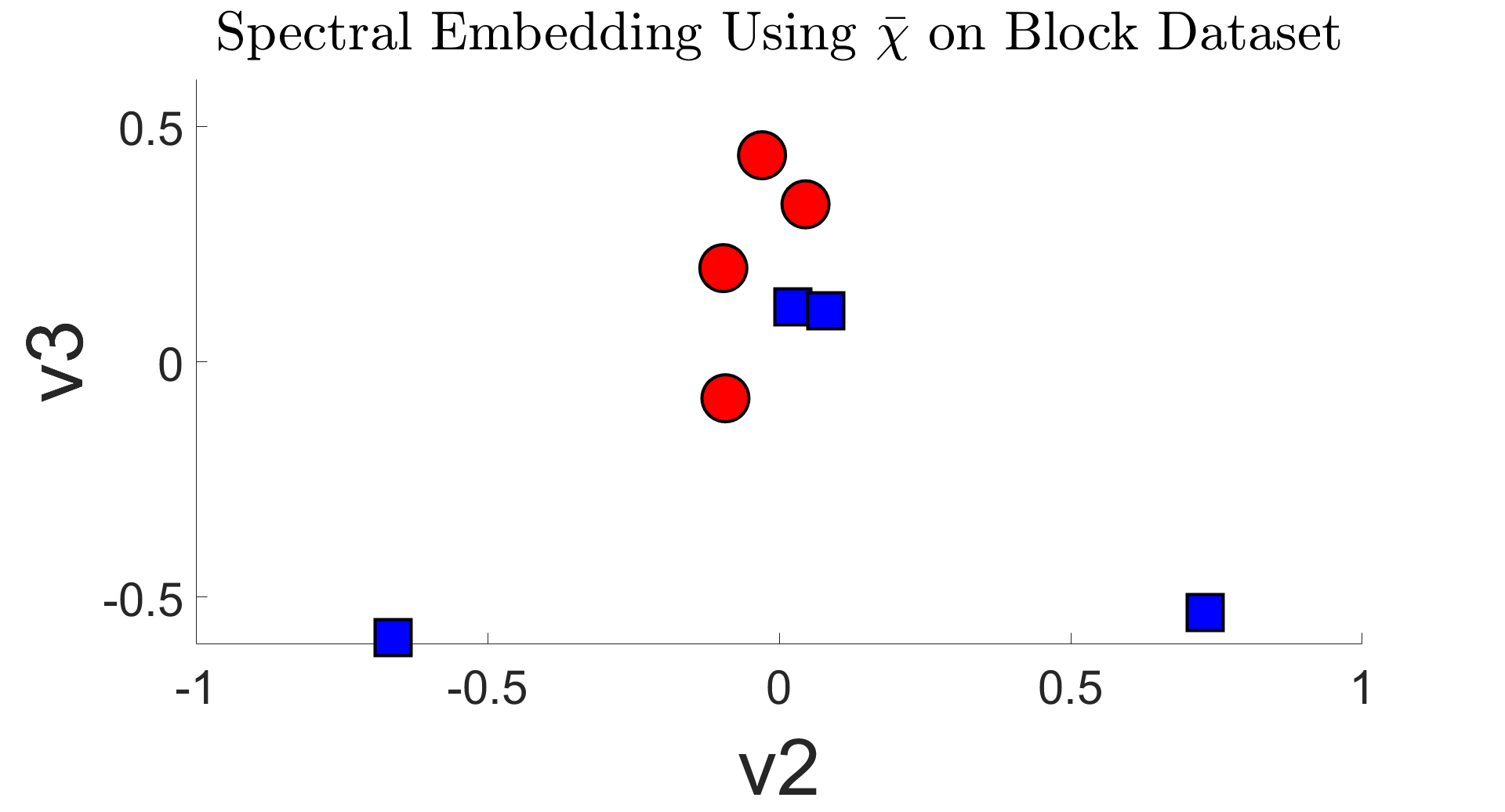}
\includegraphics[width=0.22\textwidth]{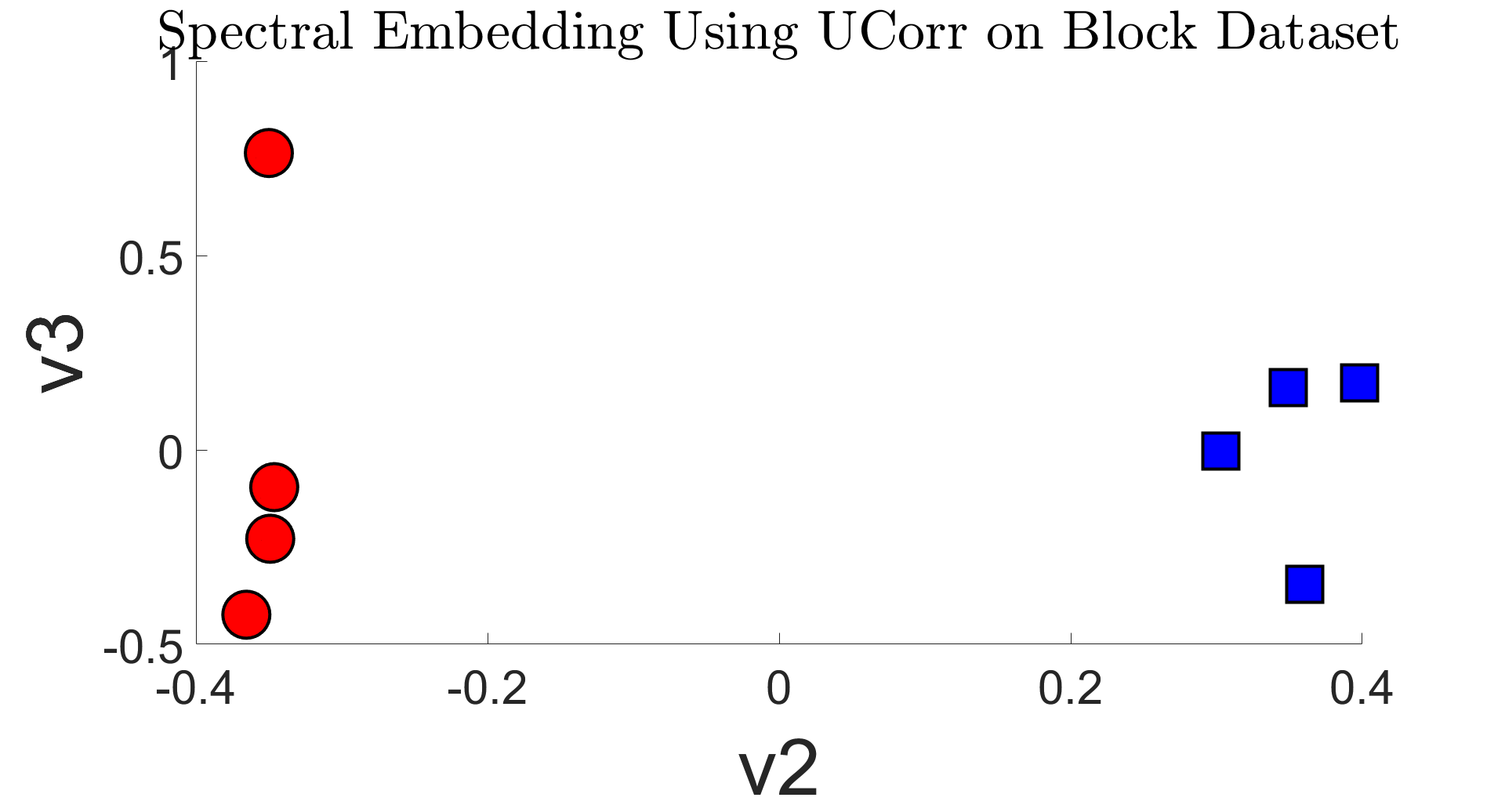}
\caption{\label{fig:BlockEmbed} By construction of this dataset the first four dimensions (red) share a tight upper block, as do the last four (blue). Thus we expect a clustering of these first four and last four. However only our method $\ta{}$ and UCorr are able to pick up on this clustering, both $\ch{}$ and $\cbh{}$ fail due to their insensitivity to location in the upper block.}
\end{figure}
%
%
\subsection{Mixture Dataset}
\begin{figure}
\centering
\includegraphics[width=0.15\textwidth]{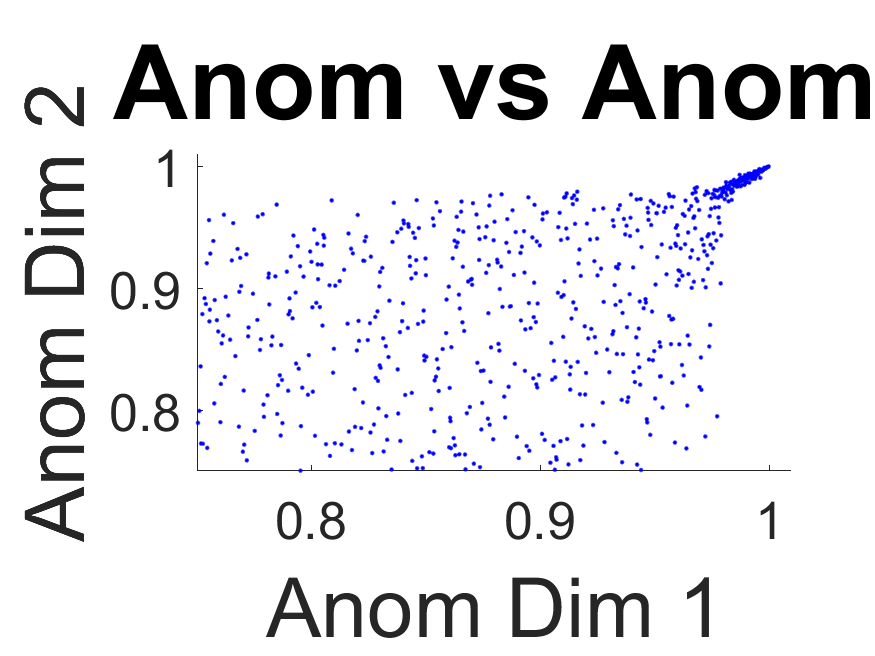}
\includegraphics[width=0.15\textwidth]{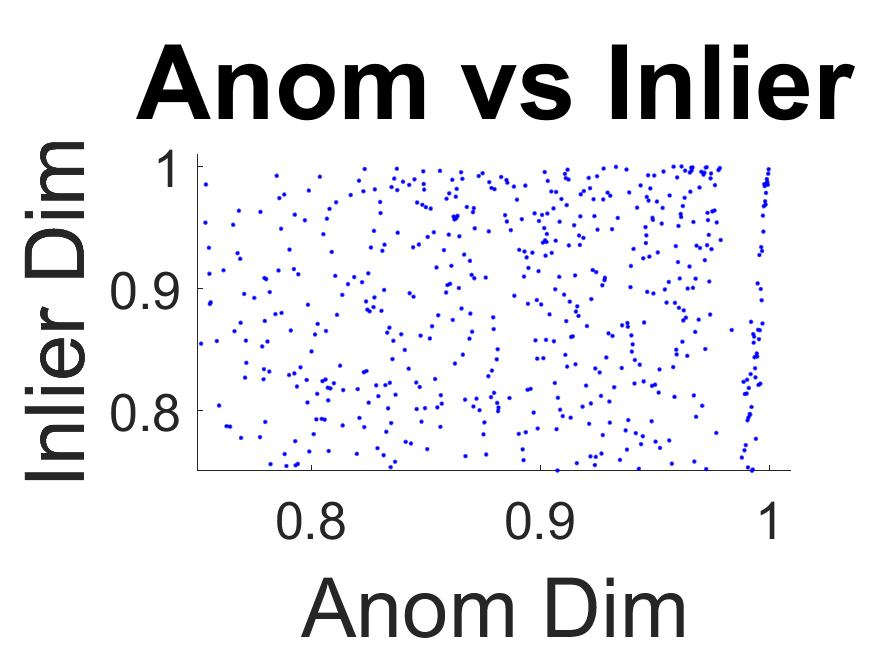}
\includegraphics[width=0.15\textwidth]{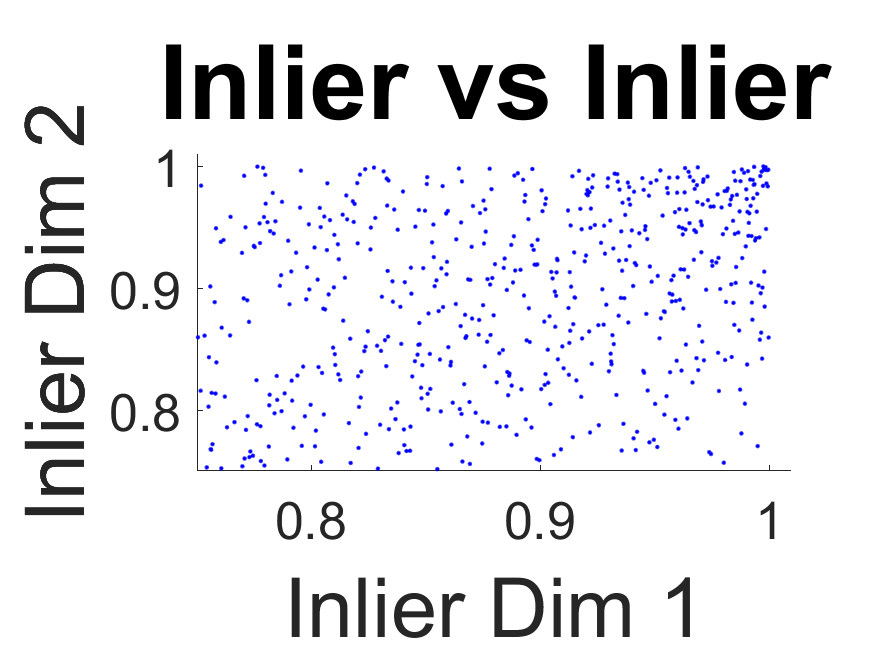}
\caption{\label{fig:MixData} Scatterplots of the Mixture Dataset for the three types of pairs (zoomed in to upper quadrant). The spiked-spiked pairs have strong tail dependence, whereas the non-spiked pairs have mild tail dependence due to correlation of the inlier component. The spiked vs non-spiked pairs have the weakest tail dependence since large values in the anomaly dimension will correspond to the anomalies, which do not have large values on the non-spiked dimensions.}
\end{figure}

\begin{figure}
\centering
\includegraphics[width=0.22\textwidth]{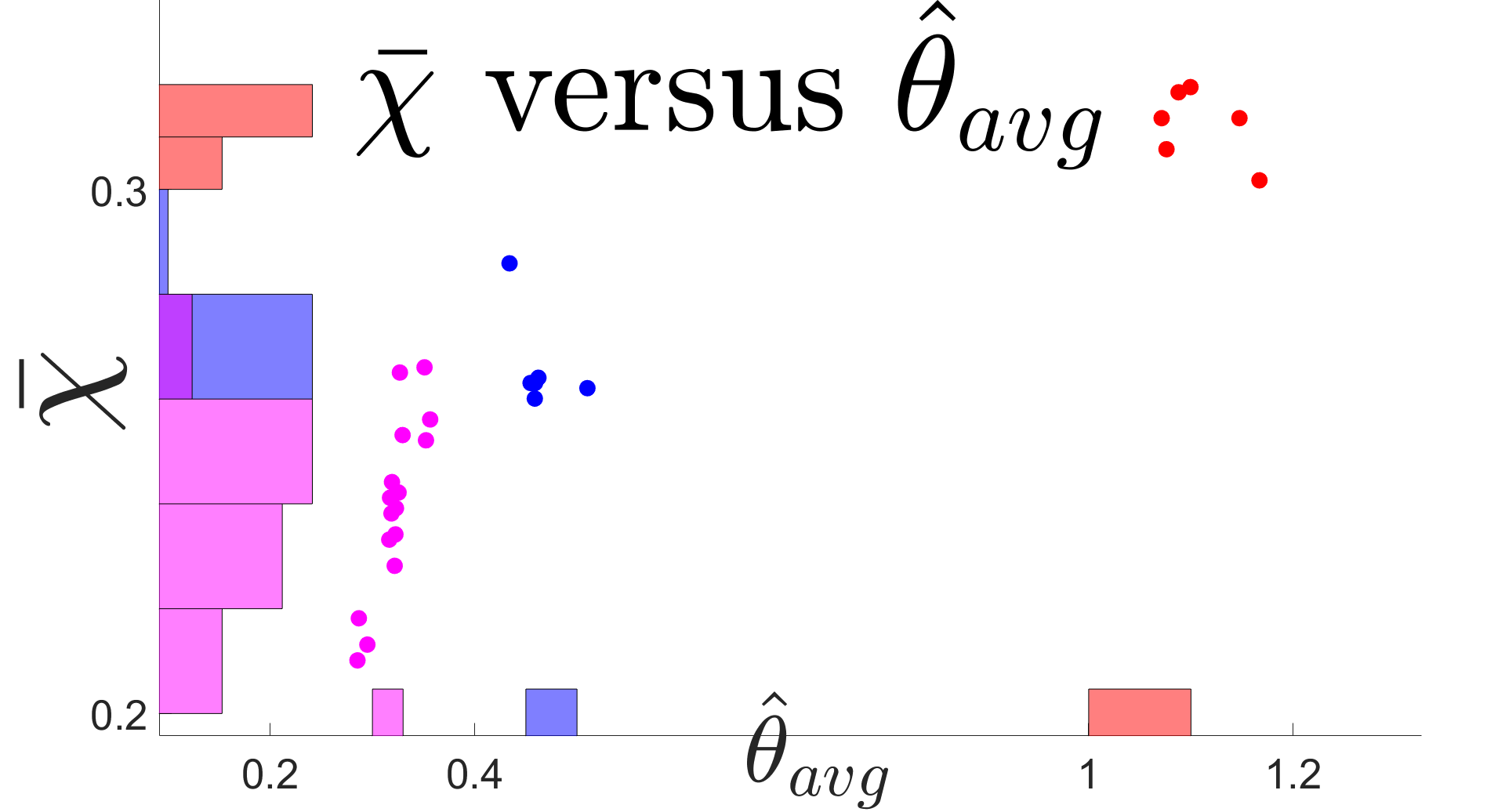}
\includegraphics[width=0.22\textwidth]{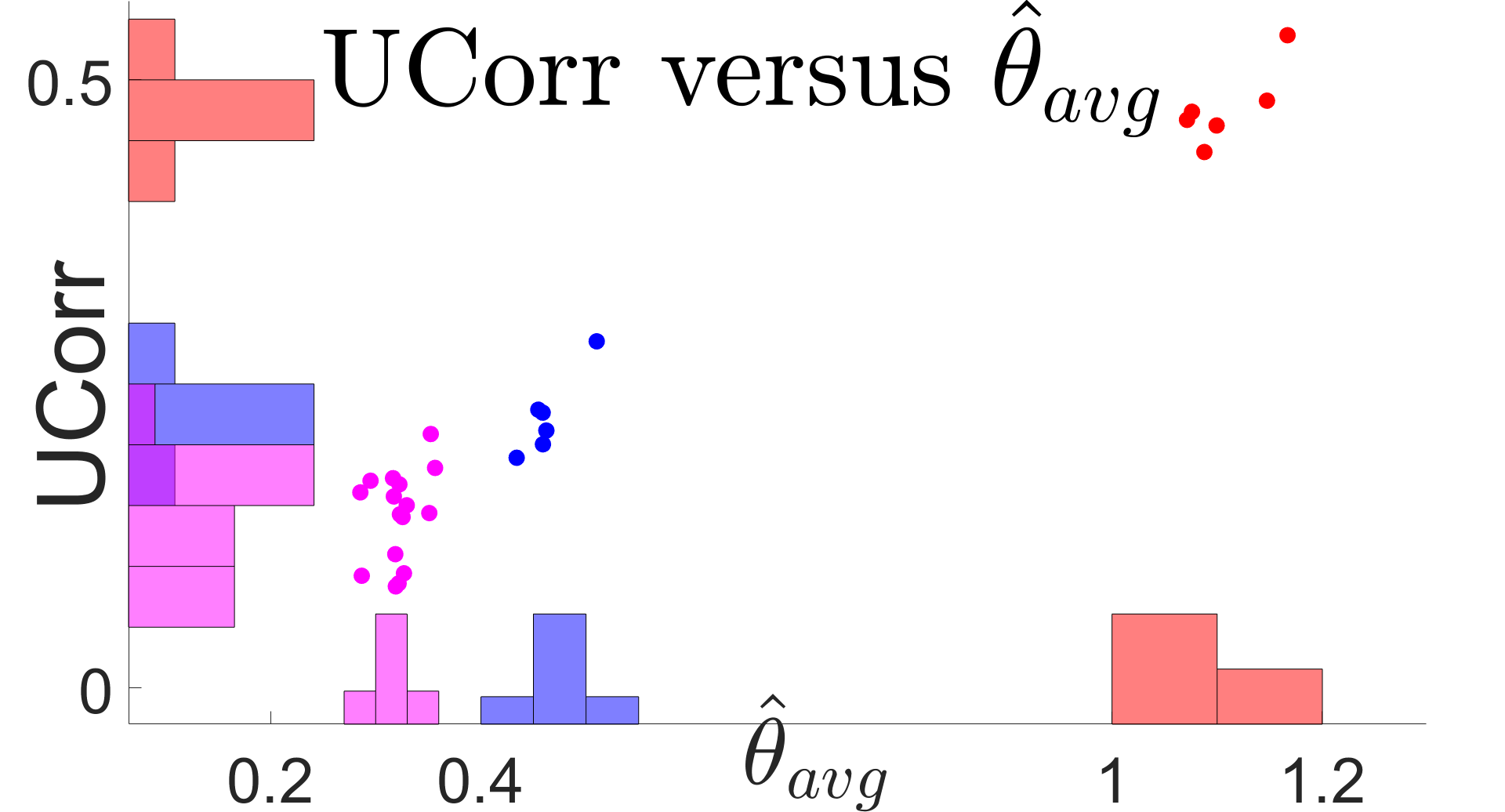}
\caption{\label{fig:MixtureScatHist} Scatterplot and Histgorams on Mixture Model dataset, comparing the proposed model to $\bar{\chi}$ (left) and  UCorr (right). We seek a similarity measure that can separate the three types of pairs, red is spiked-spiked, blue is normal-normal, and purple and spiked-normal. We see our proposed method $\ta{}(q)$ on the x-axis of both plots is able to separate the three clusters, with significant separation for the anomaly-anomaly cluster. This is because of the strong tail dependence property of the Clayton copula which captures the similarity between the anomalous dimensions, even though the anomaly fraction is small.}
\end{figure}

A scenario that is motivated by anomaly detection is data generated from a mixture, with one mixture component considered anomalous, somewhat similar to copula mixtures defined by \cite{tewari2011parametric}. The full description of this constructed dataset in is in the appendix, but in short the anomalous component is created by spiking only a subset of features. Pairs of these spiked features will possess stronger tail dependence due to these anomalies, whereas a pair of features containing one spiked dimension and one non-spiked dimension will lack tail dependence due to the lack of extreme anomaly features in the non-spiked dimension. The third type of pair, non-spiked vs non-spiked will have moderate tail dependence due to the general correlation of the non-anomalous component chosen for the dataset. These three types of pairs are shown in figure \ref{fig:MixData}.
%
We show three different similarity measures, our $\ta{}(q)$, $\cbh{}$, and UCorr, all with the value $ q = 0.75$ in figure \ref{fig:MixtureScatHist}. A good similarity measure will assign significantly higher values of similarity of anomalous dimensions with other anomalous dimensions, which are those pairs show in red. We see that our proposed $\ta{}(q)$ is the similarity measure that best separates these pairs, and is the only one that separates all three types of pairs. Our similarity measure is able to capture the tail dependence of the anomalous dimensions, although the anomalies are only a small fraction of the dataset. This was achieved due to the Clayton copula's sensitivity to tail dependence. Results from the mixture dataset are also shown in table \ref{table:Main Results}.
\subsection{Two Anomaly Modes Dataset}
\begin{figure}
\centering
\includegraphics[width=0.15\textwidth]{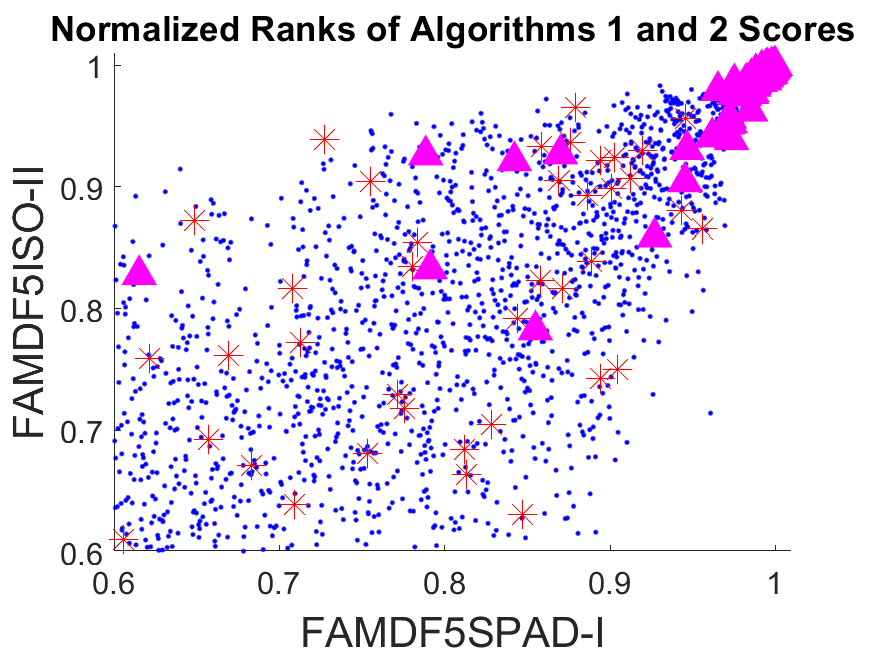}
\includegraphics[width=0.15\textwidth]{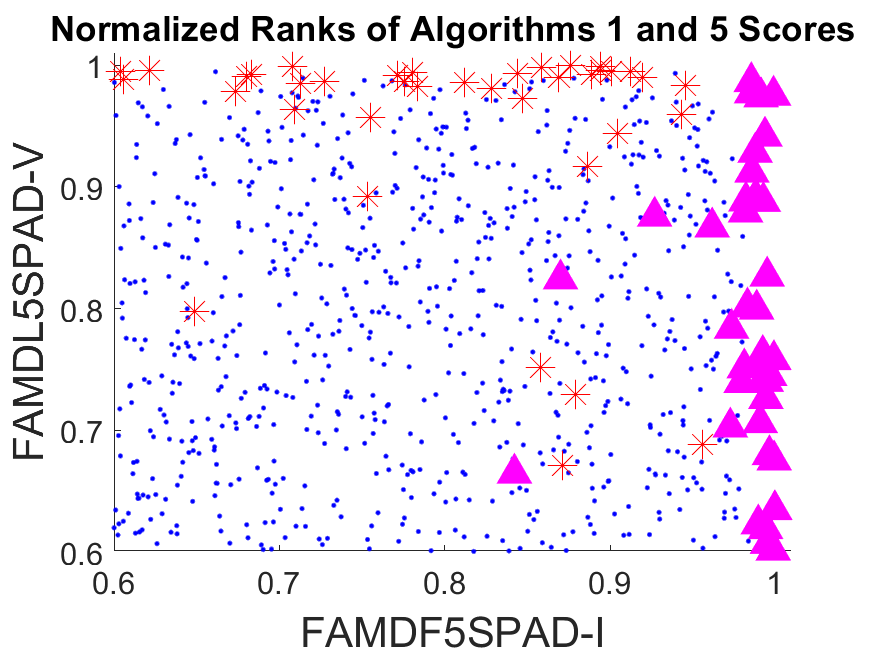}
\includegraphics[width=0.15\textwidth]{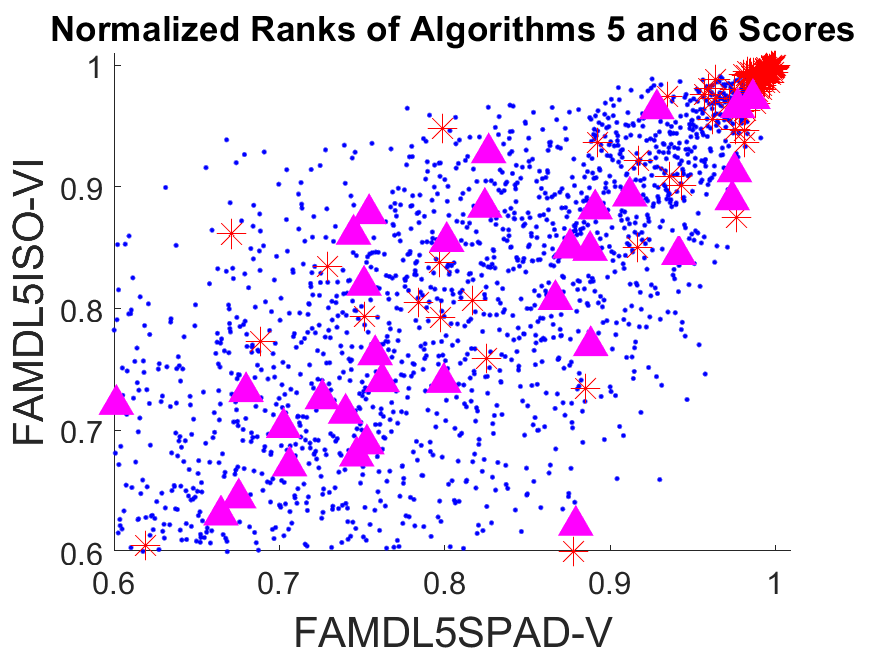}
\caption{\label{fig:TwoAnomRankScale} An example of the three types of pairs of the Two Anom dataset. The first pair is of algorithms that both use the first principal coordinates, both of these algorithms detect the purple set of anomalies well. The second pair contains one algorithm that uses the first coordinates, and the other algorithm uses the last coordinates. We see they do not agree on the anomalies, and there is minimal dependence and negligible tail dependence. The last pair is of two algorithms that both use the last set of principal coordinates, both detect the same set of red anomalies, creating similar tail dependence as the first pair shown.}
\end{figure}

We validate the proposed similarity measure to evaluate the similarity between anomaly detection algorithms on an ideal simulated dataset, which we name the TwoAnom dataset. This dataset is useful to analyze because it has two kinds of anomalies, whose anomalous behavior lie in two separate subspaces, see the appendix for the full description. In brief, for FAMD/PCA like algorithms, one kind of anomaly is revealed in the subspace containing the first few dimensions (first few principal coordinates), whereas the subspace containing the last few dimensions perform this separation for the second kind of anomaly. The anomaly scores for three example pairs are shown in figure \ref{fig:TwoAnomRankScale}. We expect a clustering of algorithms that use the first few dimensions, and a second cluster of algorithms that use the last few dimensions. We use these two clusters as ground truth and measure the Davies-Bouldin index across a variety of similarity measures, whose results are shown in table \ref{table:Main Results}.
%
%
%
\subsection{Experiments}
We perform the full flow of the proposed methodology, as shown in figure \ref{fig:copFlow}, on several datasets. One is from the UCI machine learning repository \citep{Dua:2019}, and three from the ODDS anomaly detection repository 
Descriptions of these datasets are presented in the appendix.  We apply PCA-like anomaly detection algorithms, where some of the algorithms make use of kurtosis, which has been shown to increase anomaly detection performance across several domains, see the appendix for more explanation of these algorithms \citep{davidow2020factor}. 
We expect the kurtosis methods to be more accurate and similar to each other, as they extract similar features. Thus we expect a clustering of kurtosis methods, and measure the DB index based on this expected clustering, whose results are shown in table \ref{table:Main Results}.
\begin{table}
\caption{Main Results Table $q = 0.75$}
\label{table:Main Results}
\begin{tabular}{lllll}
Dataset  & $\ta{}$ & UCorr  & $\bar{\chi}$ & $\chi$ \\
Mnist    & 0.072         & 0.075 & 0.313  & 0.321  \\
Ann      & 0.063        & 0.114 & 0.111  & 0.114  \\
Two Anom & 0.001         & 0.037 & 0.004  & 0.005  \\
Mixture  & 0.022         & 0.023 & 0.263  & 0.265  \\
Block    & 0.016         & 0.047 & 4.804  & 4.838  \\
Musk     & 0.762         & 1.059 & 1.015  & 1.023  \\
Shuttle  & 0.611         & 1.086 & 0.841  & 0.842 
\end{tabular}
\end{table}
\section{Anomaly Score Combination}
We illustrate how our similarity measure can be used to optimally combine anomalies scores from various methods in an unsupervised fashion \cite{aggarwal2017outlier,aggarwal2013outlier}. A principal component based unsupervised ensemble regression method is presented in \cite{dror2017unsupervised}, however it is assumed predictors make independent errors, which is not a valid assumption in most realistic settings. A spectral clustering approach for an ensemble of unsupervised classifiers is shown in \cite{jaffe2016unsupervised}. We present a similar method, except we have continuous anomaly scores, and these scores are on different scales. Thus we first put the anomaly scores on the same scale via equation \ref{eq:empCDF}. As we care most about the anomalies, i.e. the observations with the most extreme scores, we use our tail sensitive similarity measure to effectively cluster the scoring methods. This allows us to obtain a stronger signal within cluster, and to ignore methods which are not similar to any others, which are assumed to be noise.

We simulate a dataset by drawing from a t-distribution, chosen with a low number of degrees of freedom $\nu = 2.1$. This is chosen due to the heavy tailed property of this distribution, which will induce outliers which are many standard deviations from the mean. We note that a random variable $X$ is $t$-distributed with $\nu$ degrees of freedom if $X = Z\sqrt{\frac{\nu}{V}}$ where $Z$ is a standard normal, and $V$ is independent of $Z$ and chi-square distributed with $\nu$ degrees of freedom \citep{ruppert2011statistics}.

We partition variables into clusters and create dependence of variables within clusters by sharing the same $V$ within clusters. That is we let $j^{\text{th}}$ dimension of observation $i$ be: $X_{ij} = Z_{ij}\sqrt{\frac{\nu}{V_{ic(j)}}}$, where  $c(j)$ is the cluster index associated with $j$. We create $4$ such clusters, each with three variables. I.e. $[c(1),c(2),c(3),c(4),\ldots,c(12)] = [1,1,1,2,\ldots,4]$  We also create $50$ ``noise" dimensions, which are chosen to be normally distributed instead of t-distributed. That is for $j > 12$, we define $X_{ij} = Z_{ij}$. We flag an observation as an outlier if its distance to the origin when projected to the subspace of the first 12 dimensions is among the top $3\%$ among all all observations. To simulate the scores of anomaly detection algorithms, we consider one algorithm per dimension of the dataset, whose score on observation $i$ is the absolute value of $X_{ij}$. That is we create 62 anomaly detectors, one for each dimension, where the $j^{\text{th}}$ anomaly detector uses the $j^{\text{th}}$ feature, $\mathcal{A}_j(X_i) = |X_{ij}|$

To effectively recover the ground truth clusters, a measure is required that is sensitive to tail dependence. We compute $\ta{}$ on all pairs of anomaly detection algorithms, forming the similarity matrix $W$, where $w_{ij}$ is the similarity defined by $\ta{}$ applied to the anomaly scores of $\mathcal{A}_i$,$\mathcal{A}_j$. We transform this pairwise similarity matrix to a pairwise dissimilarity matrix with $D:=e^{-W}$ (exponential taken componentwise). We then employ the DBSCAN \citep{ester1996density} clustering algorithm on this dissimilarity matrix. We find we can exactly recover the ground truth clusters.

We create an ensemble method by taking the mean rank within cluster, and the maximum rank across these means. This allows us to produce a stronger signal within cluster, while picking up different anomalies picked out by different clusters by choosing the max function \citep{aggarwal2015outlier}. This is compared to taking the mean rank across all algorithms, and the max rank across all algorithms. We objectively measure the performance of these ensemble methods by computing the Area Under the Curve of the Receving Operating Characteristic (AUC ROC), which is the area under the curve of true positives rate versus false positive rate. The AUC ROC can be interpreted as the probability a random outlier is assigned a higher score than a random inlier.

\begin{table}
    \centering
     \begin{tabular}{c|c|c|c}
        Combine Across/Within & All & Mean & Max \\
        Mean & 0.692 & 0.864 & 0.684 \\
        Max & 0.864 & 0.999 & 0.875 
    \end{tabular}
    \caption{AUC ROC of different ensemble approaches. For the column headers ``All" refers to using all of the scores, whereas ``Mean" refers to aggregrating the scores within cluster by the mean of the score within cluster.}
    \label{tab:Ensemble}
\end{table}
\section{Discussion}
Our model outperforms $\ch{}$,$\cbh{}$, and UCorr because the survival Clayton copula puts heavy mass in the upper right tail. The similarity measure using the upper quadrant of the survival Clayton copula has been demonstrated to be sensitive to tail dependence, which is often induced only by a small fraction of extreme anomalous observations. By constructing such a similarity measure that is sensitive to this small but important fraction, one can better solve downstream tasks, such as clustering the algorithms together, recognizing risk, and ultimately combining anomaly scoring algorithms together optimally.
%
%
%

\textbf{Comparison with $\ch{}$ and $\cbh{}$.}
$\cbh{}(q)$ at a fixed level of $q$ is simply a function the fraction of points inside of $R_q$. All points inside $R_q$ are counted equally for $\cbh{}(q)$, despite the important additional information such as if they are in the top right versus the other corners. This is similar to our $\tf{}$. However our full approach using the average of both $\tf{}$ and $\ts{}$ incorporates the shape and tail dependence strength information by making use of Clayton copula's density function, assigning higher likelihood to points in the top right corner. It is for this reason that our method is able to better suited to measure similarity across a wide range of $q$, even at values far lower than the anomaly percentage, which may be unknown.
%

\textbf{Comparison with UCorr.}
The density of the Clayton copula is ideal to measure similarity that is sensitive to extreme value anomalies. Using UCorr is somewhat similar to our $\ts{}$ measure, however $\ts{}$ better captures upper tail dependence. In particular, UCorr is symmetric and does not focus on the upper tails. Further, correlation is not as sensitive to strong tail dependence and instead measures central tendencies, as compared with our highly sensitive proposed measure $\ta{}(q)$. 
%

The similarity measure using the upper right quadrant of copulas is both novel and excellently sensitive to extreme value dependence. The survival Clayton copula is well suited for this context because of its large extreme value dependence, performing significantly better than the Gaussian copula. Upper correlation has inadequate sensitivity to tail dependency. $\cbh{}(q)$ and $\ch{}(q)$ at fixed $q$ capture information only about the number of points in $R_q$, which is also captured by our proposed $\tf{}(q)$. However our method includes the more complex $\ts{}(q)$, which when averaged with $\tf{}(q)$ creates a robust measure of tail dependence.

\pagebreak
\nocite{*}

\onecolumn
\begin{centering}
\textbf{Appendix}
\end{centering}
\section{Davies Bouldin Index}
We termed the spectral embedding space $\mathbf{V}$, thus $v_{jk}$ is the $k^{\text{th}}$ coordinate of observation $j$ after spectral embedding. As all of the examples considered consisted of only two clusters, we used a one dimensional spectral embedding. However this can be modified to higher dimensional embeddings for different problems with more than two clusters. This one dimensional embedding is $\mathbf{V_2}$, it is the second column of $\mathbf{V}$, the first column of $\mathbf{V}$ contains no useful information as it corresponds to the zero eigenvalue, and is proportional to the vector of all ones. Now we present the necessary definitions for the Davies-Bouldin Index.

For each cluster $i$ the inter-cluster spread is defined as $s_i:= \frac{1}{n_i}(\sum_{j \in G(i)} ||{v}_{j2} - {a_i}||_2^2)^{1/2}$, where $a_i$ is the centroid of cluster $i$,and $G(i)$ is the set associated with cluster $i$. The intra cluster distance $m(i,j)$ is defined between cluster centers , $m_{ij}:= ||a_i - a_j||_2$. From this a pairwise loss function is defined between clusters $r_{ij} := \frac{s_i + s_j}{m_{ij}}$, and the loss of a single cluster is defined as $d_{i} = \text{max}_{j}(r_{ij})$. The Davies-Bouldin index as defined as $ \text{DB}:= \frac{1}{n_c} \sum_{i=1}^{n_c} d(i).$
Where in our examples $n_c$ = 2 is the number of clusters. For clusters than are well separated, $s_i$ is small compared to $m_{ij}$, and thus the Davies-Bouldin index is small.


\section{Data Description}
\subsection{Mixture Model}
Tewari (2011) defines a copula generated from a Gaussian Mixture Model. We use a slightly different model; there are two mixture components, inliers generated from a $d = 8$ dimensional multivariate normal with pairwise correlation $\rho_1 = 0.6$, and anomalies generated with pairwise correlation $\rho_2 = 0.8$, whose first $d_a = 4$ dimensions are scaled by a factor $c = 10 $. The reason for spiking only the first $d_a$ dimensions is to create a clustering of columns, the first $d_a$ columns have similar extreme points (the anomalies), whereas the rest of the columns have more independent extreme points. Absolute values are taken of the dataset so the anomalous behavior lies only in the upper tail. The data set is generated by drawing 5000 inliers and 200 anomalies from the above mentioned multivariate Gaussians, taking the absolute value of all features, then taking the empirical CDF transform, equation 10 in the main paper so the data is on the copula scale.  There are three kinds of pairs generated from such a model, an anomalous column with another anomalous column, an anomalous column with a regular, and a regular with a regular. 

\subsection{Two Anom Dataset Description}
We aim to create a dataset with two different kinds of anomalies picked up in two different subspaces. We create a dataset with latent dimension 100, where the main variation of the inliers is only among a 30 dimensional ``true" subspace, variation on the remaining 70 dimensional subspace is a factor of 100 smaller. One set of anomalies has the same covariance structure on this 30 dimensional subspace, but more noise (larger magnitude by a factor of 30) on the remaining 70 dimensional subspace as compared to the inliers. The second set of inliers is ``spiked" (larger magnitude by a factor of 3) on a 5 dimensional subspace of the 30  dimensional ``true" subspace, with identical structure to the inliers on the remaining 95 dimensional subspace. We create such a dataset with 5000 inliers, and 200 total anomalies, 100 of each of these two types. The first type of anomaly is separated by the smallest principal components, as those are the noise components of inliers which these anomalies differ on. The second type of anomaly is separated by the largest principal components, since these anomalies contribute to these components making them the largest. We use a set of four anomaly detection algorithms that make use the first principal components, and a set of four anomaly detection algorithms that make use of the last principal components, as described in (Davidow, 2020).

\subsection{Experimental Datasets}
We use an anomaly detection dataset from the UCI machine learning repository, called the ann dataset.   We use three different anomaly detection from the ODDS dataset, Mnist, Musk and Shuttle. More detailed descriptions of those are found at the ODDS site (Ray). We show summary statistics of these datsets in table \ref{table:dataSummary}.
%

\begin{table}[ht]
\begin{center}
\caption{Summary Statistics of Experimental Datasets}
\label{table:dataSummary}

\begin{tabular}{llllll}
Dataset  & Observations & Dimensions & Outliers(\%) \\
Mnist    & 7603 &  100 &	700 (9.2\%) \\ 
Ann      & 7200 &   21 &     533(7.4\%) \\   
Musk     & 3062 & 166  &	97 (3.2\%) \\
Shuttle  & 49097 & 	9	& 3511 (7.0\%) 
\end{tabular}
\end{center}
\end{table}

 We apply PCA-like anomaly detection algorithms, where some of the algorithms make use of kurtosis, which has been shown to be more accurate and similar to each other, as they extract similar features which has been shown to increase anomaly detection performance across several domains (Davidow, 2020).

\subsection{Correlation between $\tf{}$ and $\ts{}$}
The claim was made in the original paper that the correlation between $\tf{}$ and $\ts{}$ is small when the survival Clayton model is well specified. We motivate this numerically by drawing survival clayton samples with three different values of $n= 100,500,1000$ with $q=0.75$. For each sample $\tf{}$ and $\ts{}$ are computed. The sample correlation of $\tf{}$ and $\ts{}$ are computed for each $n$
using $10000$ complete redraws of the $n$ samples. The resulting sample correlations of these $10000$ redraws were $-0.0318,-0.032,-0.0014$ for $n=100,500,1000$, respectively. These two estimators have small bias, and also do not have a large positive correlation, and thus a weighted combination of them will produce a superior estimator.
\vfill

We prove that $\tf{}$ and $\ts{}$ are both unbiased up to order $O(1/n)$.

The setting is that $n$ independent samples $\lbrace \mathbf{u_i} \rbrace_{i=1}^n$ are drawn with the survival Clayton distribution, equation 7 in the main paper, with true parameter $\theta_0$.
We begin by proving that $\tf{}$ has bias $O(1/n)$.
As a reminder we defined $N_q = \# \lbrace \mathbf{u}_i \in R_q \rbrace$ with $R_q = [q,1]^2$. Since each $\mathbf{u}_i$ is iid, $N_q$ is binomially distributed, with $n$ trials and success parameter  
 $p := P (\mathbf{u_1} \in R_q ) = S(q,q|\theta_0)$.
Recall $\hat{S} = N_q/n$, and recall that $\tf{}$ can be solved by equation 14 in the main paper, thus if we denote $g(\theta) = S(q,q|\theta)$ then we see conditional on $N_q, \tf{}$ is a deterministic function of $\hat{S}$: 
$$h(\hat{S}): = \tf{} = g^{-1}(\hat{S})$$. 

We know $g$ has an inverse because $S(q,q|\theta)$ is strictly monotonic in $\theta$. However there are boundary issues when $\hat{S} < (1-q)^2$ and when $\hat{S} \geq (1-q)$, for these we define $h(\hat{S})$ as $0$ and $\theta_M$ respectively for some arbitrary but finite $\theta_M > 0$. In order for $\tf{}$ and $\ts{}$ to have $O(1/n)$ bias the true $\theta_0$ must satisfy $\theta_0 \in (0,\theta_M)$.

Thus $E(\tf{}) = E(h(\hat{S}))$. At any fixed $0 < q < 1, S(q,q|\theta)$ is $C^\infty$  w.r.t $\theta$, so $h(\cdot)$ is also $C^\infty$ by the inverse function theorem, at least for $ (1-q)^2 < \hat{S} < (1-q)$ (i.e. at least in some $\delta$ interval around $ S(q,q|\theta_0)$). Thus we can appeal to the results
of Khan (2004), to conclude 
$$E(h(\hat{S})) = h(p) + \frac{\sigma^2}{2n}h''(p) + O(1/n^2),$$

where $\sigma^2 = p(1-p)$ is the variance of the indicator variable $\lbrace \mathbf{u}_i \in R_q \rbrace$. $h''(p)$ is independent of $n$, and $h(p)$ is the true $\theta_0$ since $h(p) = g^{-1}(p) = g^{-1}(S(q,q|\theta_0))$. Thus, the bias is $O(1/n)$.

Now we turn our attention to $\ts{}$, and prove that it is unbiased up to $O(1/n)$. We again put a bound on the estimator, the space over which $\theta$ is optimizied is on $[0,\theta_M]$, and this estimator is unbiased up to $O(1/n)$ only when the true $\theta_0 \in (0,\theta_M)$. We also let $\ts{}$ take the value $0$ when $N_q$ = 0, although this will happen with exponentially small probability in $n$.

Of critical importance is that the definition of $\ts{}$ coincides with that of the maximum likelihood estimator for the distribution in equation 8 from the main paper. Thus when there are $n_q$ points inside of $R_q$, $\ts{}$ is unbiased up to order $O(1/n_q)$ That is
$$E(\ts{}|N_q = n_q ) = \theta_0 + r(n_q).$$

Where $r(n_q)$ is the bias of the MLE when there are $n_q$ data, in particular $r(n_q)$ is $O(1/n_q)$ since $\ts{}$ is a maximum likelihood estimator, and $r(n_q)$ is bounded since we have defined a bounded $\ts{}$.

\begin{equation*}
\begin{split} &E(\ts{}) = \sum_{n_q = 0}^n E(\ts{}|N_q = n_q) P(N_q  = n_q)\\
&= \theta_0 + \sum_{n_q = 0}^n r(n_q)P(N_q = n_q)\\
&= \theta_0 + \sum_{n_q = 0}^{\nf{}} r(n_q) P(N_q  = n_q) + \sum_{n_q = \nf{} + 1}^n r(n_q) P(N_q  = n_q)  
\end{split}
\end{equation*}

We first argue the first summation is less than $O(1/n)$. We use the central limit theorem approximation to the binomially distributed $N_q$, which has mean $np$ and variance $np(1-p)$. The upper bound of the sum is $O(\sqrt{n})$ standard deviations from the mean, and thus appealing to the normals cdf asymptotics, $\sum_{n_q = 0}^{\nf{}}  P(N_q  = n_q)$ is $O(e^{-n}/\sqrt{n})$, which in particular is $O(1/n)$.

The second summation is $O(1/n)$ since $r(n)$ is $O(1/n)$. To spell it out, we have for some $c > 0$, $r(n) < c/n$. For notational simplicity we let $b = \nf{} + 1$. As $1/n$ is a decreasing function, $\forall n \geq b, r(n) < c/b$. Using this we can bound 
$$\sum_{n_q = b}^n r(n_q) P(N_q  = n_q) < \sum_{n_q = b}^n \frac{c}{b} P(N_q  = n_q) < \frac{c}{b}\sum_{n_q = b}^n P(N_q  = n_q) < c/b.$$ This last bound $c/b$ is $O(1/n)$ (since $b = \nf{} + 1$), thus the bias of $\ts{}$ is also $O(1/n)$.

\subsection{Asymptotic Independence}
In fact we can prove along similar lines that $\ts{}$ and $\tf{}$ are asymptotically independent. To make this precise, we define the shifted and scaled estimators as indexed by $n$:
$\tsn{} = \sqrt{n}(\ts{} - \theta_0)/\sigma_s$ and 
$\tfn{} = \sqrt{n}(\tf{} - \theta_0)/\sigma_f$.

The results of Khan (2004) also prove that the variance of $\tf{}$ decreases with $n$, thus $\tfn{}$ converges in distribution to $N(0,1)$ for correctly chosen $\sigma_f$.

We let $F^{n_q}(x): = P(\tsn{} \leq x| N_q = n_q)$. We rely heavily on the fact that the definition of $\ts{}$ coincides with the definition of the maximum likelihood estimator of $n_q$ data drawn independently inside of $R_q$ according to equation (8) in the main paper. Thus $F^n(x)$ corresponds to the cdf of a shifted and scaled maximum likelihood estimator, and thus with the correct choice of $\sigma_s$, $F^n(x) \rightarrow \Phi(x)$, where $\Phi(x)$ is the standard normal's cdf.

\textbf{Theorem}:  $\tsn{}$ and $\tfn{}$ are asymptotically independent 

\textbf{Proof} 
The definition of asymptotic independence is that the multivariate distribution of the pair of random sequences converges to the joint distribution of independent random variables. Thus we aim to prove the following convergence statement: 
$$\forall x,y, P(\tsn{} \leq x,\tfn{} \leq y) \rightarrow \Phi_2(x,y),$$

as $n \rightarrow \infty$, where $\Phi_2$ is the two dimensional multivariate centered normal cumulative distribution function whose covariance matrix is the 2 by 2 identity matrix. 

We need to show that $\forall \epsilon > 0 $ and $\forall x,y \in \mathbb{R}$, that $\exists N$ such that $\forall n\geq N, | P(\tsn{} \leq x,\tfn{} \leq y) - \Phi_2(x,y)| < \epsilon$
By the fact that $\tsn{}$ and $\tfn{}$ both converge in distribution to a standard normal, we know $\forall \epsilon_2,\epsilon_3 > 0$ and $\forall x,y \in \mathbb{R},\exists N_2$ such that $\forall n \geq N_2$, $|F^n(x) - \Phi(x)| < \epsilon_2$, and $|P(\tfn{} \leq y) - \Phi(y)| < \epsilon_3 $. We choose $N$ at least large enough so that ${\floor{Np/2}} \geq N_2$, e.g. $N \geq \ceil{2N_2/p}$. We denote $n_p:= \floor{Np/2}$ for notational simplicity. We proceed along the same lines as above to compute the probability of interest, $P(\tsn{} \leq x,\tfn{} \leq y)$, by conditioning on possible values of $N_q$. Given the condition $\tfn{} \leq y$, the largest relevant value of $N_q$ is $n_b:= \floor{N\cdot g(\frac{y\sigma_f}{\sqrt{N}} + \theta_0)}$, obtained by solving $n_q$ from the equation defining $\tfn{}$, i.e. $n_b:= \text{argmin} \lbrace n_q | \tfn{}(n_q) <= y \rbrace $. Thus,

\begin{align}
    P(\tsn{} \leq x,\tfn{} \leq y) & = \sum_{n_q = 0}^n P(\tsn{} \leq x,\tfn{} \leq y|N_q = n_q) P(N_q = n_q) \nonumber \\
    & = \sum_{n_q = 0}^n P(\tsn{} \leq x|N_q = n_q)P(\tfn{} \leq y | N_q = n_q) P(N_q = n_q)  \label{l2} \\
    & = \sum_{n_q = 0}^{n_b} P(\tsn{} \leq x|N_q = n_q) P(N_q = n_q) \label{l3} \\
    & = \sum_{n_q = 0}^{n_b} F^{n_q}(x)P(N_q = n_q) \label{l4} \\
    & = \sum_{n_q = 0}^{n_p} F^{n_q}(x)P(N_q = n_q) \hspace{0.1cm} + \sum_{n_q = n_p + 1}^{n_b} F^{n_q}(x)P(N_q = n_q) \label{l5}
\end{align}
In \eqref{l2} we used the fact that conditional on the event $N_q$ = $n_q$, $\tfn{}$ is a deterministic monotonic function of $n_q$ (thus it is conditionally independent of $\tsn$ conditioned on $n_q$), and \eqref{l3} follows since $\tfn{} > y$ when $n_q > n_b$. In \eqref{l4} we re-wrote the conditional cdf of $\tsn{}$ using the definition of $F^{n}$. In \eqref{l5} we break this sum into the same two parts as done previously, from $0$ to $n_p$, and $(n_p + 1)$ to $n_b$. The first part from $0$ to $n_p$ is $O(1/N)$ along the same reasoning as shown in section 2.4. To be explicit, $|F^{n_q}(x)| < 1$ and so $|\sum_{n_q = 0}^{n_p} F^{n_q}(x)P(N_q = n_q)| < \sum_{n_q = 0}^{n_p} P(N_q = n_q)$, and $\sum_{n_q = 0}^{n_p} P(N_q = n_q)$ was shown to be $O(1/N)$ in section 2.4, and thus can be made arbitrarily small with a large enough choice of $N$. However we have made use of an implicit assumption that $n_p <= n_b$. This can always be made true with a large enough choice of $N$ (which importantly may depend on $y$), since choosing $N$ arbitrarily large one can make $n_b$ arbitrarily close to $Ng(\theta_0)=Np > n_p = \floor{Np/2}$.

We focus on the second part, 
$$ S(x,y):= \sum_{n_q = n_p + 1}^{n_b} F^{n_q}(x)P(N_q = n_q). $$

Note that $S$ depends on $y$ since the upper bound $n_b$ depends on $y$. Recall that for all $x$, $F^n(x)$ converges to $\Phi(x)$, that is $\forall n_q \geq n_p \geq N_2$, the difference
$|F^{n_q}(x) - \Phi(x)| < \epsilon_2$. We denote $D(y):= \sum_{n_q = n_p + 1}^{n_b} P(N_q = n_q)$. Thus, we can bound the difference of $S(x,y)$ and $\Phi(x)D(y)$ using the triangle inequality, $|S(x,y) - \Phi(x)D(y)| < \epsilon_2D(y)$. This can be further manipulated with the triangle inequality to bring us closer to our desired inequality relating $S(x,y)$ and $\Phi(x)\Phi(y)$:

\begin{equation}\label{eq:sBound}|S(x,y) - \Phi(x)\Phi(y)| < | \Phi(x)||D(y) - \Phi(y)| + |\epsilon_2D(y)| . \end{equation}

We now aim to bound $|D(y) - \Phi(y)|$, we use the fact that $\tfn{}$ is a monotonic function of $N_q$, thus $P(\tfn \leq y) = \sum_{n_q = 0}^{n_b}P(N_q = n_q).$ We have shown $\sum_{n_q = 0}^{n_p} P(N_q = n_q)$ is $O(1/N)$, thus $P(\tsn \leq y) - D(y)$ is $O(1/N)$, or informally $D(y) = P(\tsn \leq y) + O(1/N)$. Since $\tsn{}(y) \rightarrow \Phi(y)$, we know that $|P(\tsn \leq y) - \Phi(y)| <\epsilon_3$, and thus (also informally):
$$ |D(y) - \Phi(y)| < \epsilon_3 + O(1/N)$$ 

Plugging this bound for $|D(y) - \Phi(y)|$ into \eqref{eq:sBound}, we can bound the difference (by the triangle inequality):

$$|S(x,y) - \Phi(y)\Phi(x)| < |\Phi(x)\epsilon_3 | +  |\Phi(x)O(1/N)| + |\epsilon_2 \Phi(y)| +  |\epsilon_2\epsilon_3|  + |\epsilon_2O(1/N)| .$$

As $\Phi$ is bounded between $0$ and $1$, and $N$ is allowed to depend on $\epsilon$ and $x,y$, this error can be driven arbitrarily small for small enough choices of $\epsilon_2,\epsilon_3$, and a large enough choice of $N$. We recall that $S(x,y)$ is the second portion of the sum of the probability of interest $P(\tsn{} \leq x,\tsn{} \leq y)$, but the first part was shown to be $O(1/N)$, and thus the error $|P(\tsn{} \leq x,\tfn{} \leq y) - \Phi(x)\Phi(y)|$ can be driven arbitrarily small, and so the estimators $\tsn{},\tfn{}$ are asymptotically independent.

The intuition behind the asymptotic independence of $\tf{}$ and $\ts{}$ is that $\ts{}$ is estimating the ``shape" inside, and the number of points inside does not influence the shape but only how densely the shape is filled in. This intuition suggests that the two estimators may be independent (not just asymptotically), but we conjecture this is not the case, mainly due to the numerically sampled correlations cited above, which we found to interestingly be consistently negative. This dependence between the two is possibly due to a dependence of the bias of $\ts{}$ on $N_q$, it is only asymptotically unbiased, which is why the pair of estimators is only asymptotically independent.

\section{Main Results}
Due to brevity we only presented results at $q = 0.75$ in the main paper. However here we show results for $q = 0 .5, 0.75,0.9,0.95$


\begin{table}[ht]
\begin{center}
\caption{Results at Two $q$ Values. Left $q = 0.5$, Right  $q = 0.75$}
\label{table:q5075}
\begin{tabular}{llllllllll}
Dataset  & $\ta{}$ & UCorr  & $\bar{\chi}$ & $\chi$ & || & $\ta{}$ & UCorr  & $\bar{\chi}$ & $\chi$ \\
Mnist    & 0.248         & 0.264 & 0.494  & 0.507  & || & 0.072         & 0.075 & 0.313  & 0.321  \\
Ann      & 0.251         & 0.139 & 0.475  & 0.521  & || & 0.063        & 0.114 & 0.111  & 0.114  \\
Two Anom & 0.001         & 0.021 & 0.005  & 0.009  & || & 0.001         & 0.037 & 0.004  & 0.005  \\
Mixture  & 0.029         & 0.058 & 0.149  & 0.154  & ||  & 0.022         & 0.023 & 0.263  & 0.265  \\
Block    & 0.033         & 0.132 & 2.177  & 2.153  & ||  & 0.016         & 0.047 & 4.804  & 4.838  \\
Musk     & 0.804         & 0.870 & 0.966  & 1.123  & ||  & 0.762         & 1.059 & 1.015  & 1.023  \\
Shuttle  & 0.418         & 0.828 & 0.434  & 0.503  & || & 0.611         & 1.086 & 0.841  & 0.842 
\end{tabular}
\end{center}
\end{table}

\begin{table}[ht]
\begin{center}
\caption{Results at Two $q$ Values. Left $q = 0.9$, Right $q = 0.95$}
\label{table:q90}
\begin{tabular}{llllllllll}
Dataset  & $\ta{}$ & UCorr  & $\bar{\chi}$ & $\chi$ & || & $\ta{}$ & UCorr  & $\bar{\chi}$ & $\chi$ \\
Mnist    & 0.024         & 0.065 & 0.060  & 0.053  & || & 0.030         & 0.072 & 0.151  & 0.117  \\
Ann      & 0.069         & 0.104 & 0.112  & 0.100  & || & 0.061         & 0.244 & 0.071  & 0.062  \\
Two Anom & 0.000         & 0.120 & 0.014  & 0.008  & || & 0.075         & 0.575 & 0.024  & 0.007  \\
Mixture  & 0.006         & 0.040 & 0.059  & 0.054  & || & 0.003         & 0.434 & 0.039  & 0.028  \\
Block    & 0.011         & 0.058 & 0.040  & 0.039  & || & 0.009         & 0.174 & 0.041  & 0.034  \\
Musk     & 0.777         & 1.148 & 0.956  & 0.952  & || & 0.800         & 1.037 & 0.836  & 0.843  \\
Shuttle  & 0.964         & 1.199 & 1.060  & 1.059 & || & 0.913         & 6.234 & 1.199  & 1.212 
\end{tabular}
\end{center}
\end{table}

\bibliography{bibliography}
\end{document}